\documentclass[twoside]{article}

\usepackage[accepted]{aistats2022}
\usepackage{graphicx}
\usepackage{amsthm}
\usepackage{math_cmd}
\usepackage{xcolor}
\usepackage{algpseudocode}
\usepackage{algorithm}
\usepackage{breakcites}
\usepackage{pifont}
\usepackage{diagbox}
\usepackage{multirow}
\usepackage{booktabs}
\usepackage{enumitem}
\usepackage[titletoc]{appendix}
\usepackage{hyperref}

\usepackage{natbib}
\begin{document}


\runningauthor{Bai, Zhang, Li, Henao, Wang, Carin}

\twocolumn[

\aistatstitle{Collaborative Anomaly Detection}

\aistatsauthor{ Ke Bai$^{*}$ \And Aonan Zhang \And Zhizhong Li \And Ricardo Henao }

\aistatsaddress{Duke University \And  Bytedance Inc.  \And SenseTime Research \And Duke University}

\aistatsauthor{ Chong Wang$^{**}$ \And  Lawrence Carin }
\aistatsaddress{  Apple Inc. \And   KAUST}
% \aistatsaddress{ ke.bai@duke.edu}
% }
]

\begin{abstract}
In recommendation systems, items are likely to be exposed to various users and we would like to learn about the familiarity of a new user with an existing item. 
This can be formulated as an anomaly detection (AD) problem distinguishing between ``common users'' (nominal) and ``fresh users'' (anomalous). Considering the sheer volume of items and the sparsity of user-item paired data, independently applying conventional single-task detection methods on each item quickly becomes difficult, while correlations between items are ignored. 
To address this multi-task anomaly detection problem,
we propose collaborative anomaly detection (CAD) to jointly learn all tasks with an embedding encoding correlations among tasks.
We explore CAD with conditional density estimation and conditional likelihood ratio estimation.
We found that: $i$) estimating a likelihood ratio enjoys more efficient learning and yields better results than density estimation.
$ii$) It is beneficial to select a small number of tasks in advance to learn a task embedding model, and then use it to warm-start all task embeddings.
Consequently, these embeddings can capture correlations between tasks and generalize to new correlated tasks.
\end{abstract}

%Thank you!
% We first find that  We also propose a task embedding model, which can capture the correlations between tasks and generalized to new correlated. 

% in advance to learn a task embedding model. The learned embedding can capture the correlation between tasks

% in order to capture the correlation between tasks and generalize to new correlated tasks. 

% This paper focuses on an unsupervised learning model to estimate the familiarity a user has with each item using exposure data alone.

% However, independently applying conventional single-task AD methods to each item is inefficient provided a large number of items, and the scarcity of data for each item.
% We propose collaborative anomaly detection (CAD) to jointly learn all tasks with an embedding encoding correlations among tasks.
% We explore CAD with conditional density estimation and with conditional likelihood ratio estimation.
% Our study finds that: $i$) estimating a likelihood ratio enjoys more efficient learning and yields better results than density estimation; and
% $ii$) it is beneficial to select a small number of tasks in advance to learn a task embedding model, and then use it to warm-start all following tasks.
\section{Introduction}

% We focus on anomaly detection (AD) with many correlated tasks, where each task is only exposed to a fraction of the entire population.
% This setting is derived from the need of estimating the user-item freshness/familiarity in recommendation systems.
% Recent research on user behavior analysis shows that a user may feel fatigue if similar items are displayed to the user multiple times~\citep{ma2016user}.

User-item exposures constitute weakly labeled data when compared to strong labels such as clicks, converts ({\em e.g.}, user purchase items), and ratings.
For instance, a user may browse an item without buying it or people may watch a video without leaving a rating.
Though these situations generally do not indicate user preferences toward specific items or movies, the data generated by interactions {\em without action\/} is of interest because it is a reflection of user behavior. 
For example, the familiarity between the item and a user, with which
%
% But users no-action data is worth studying to analyze the users' behaviors.
%
% Exposure data can express
% With a familiarity measurement between a user and an item, 
the recommendation system could increase the recommendation diversity and thus improve the user experience~\citep{kunaver2017diversity}.
% Considering the case that Youtube always recommends similar videos in user's familiar area, introducing suitable diversity provides users more room to explore new things. 

%, even though the impressions are discounted~\citep{lee2014modeling}.
% From the items' perspective, the best strategy to achieve long-term benefit is to balance exploration and exploitation given the uncertainty of a group of users toward a given item.
% for each user group this item exposed to.
% This paper learns from exposure data to distinguish ``common users'' from ``fresh users'' for each item.
% in order to refrain from repeatedly exposing items to common users.
% This intuition happens to coincide with the inverse propensity weighting methods in recommendation systems~\citep{liang2016modeling}, where overly exposed user-item pairs are discounted in the loss function.

\begin{figure}[t]
  \centering
\scalebox{0.8}{
  \begin{minipage}[c]{0.48\textwidth}
    \includegraphics[width=\textwidth]{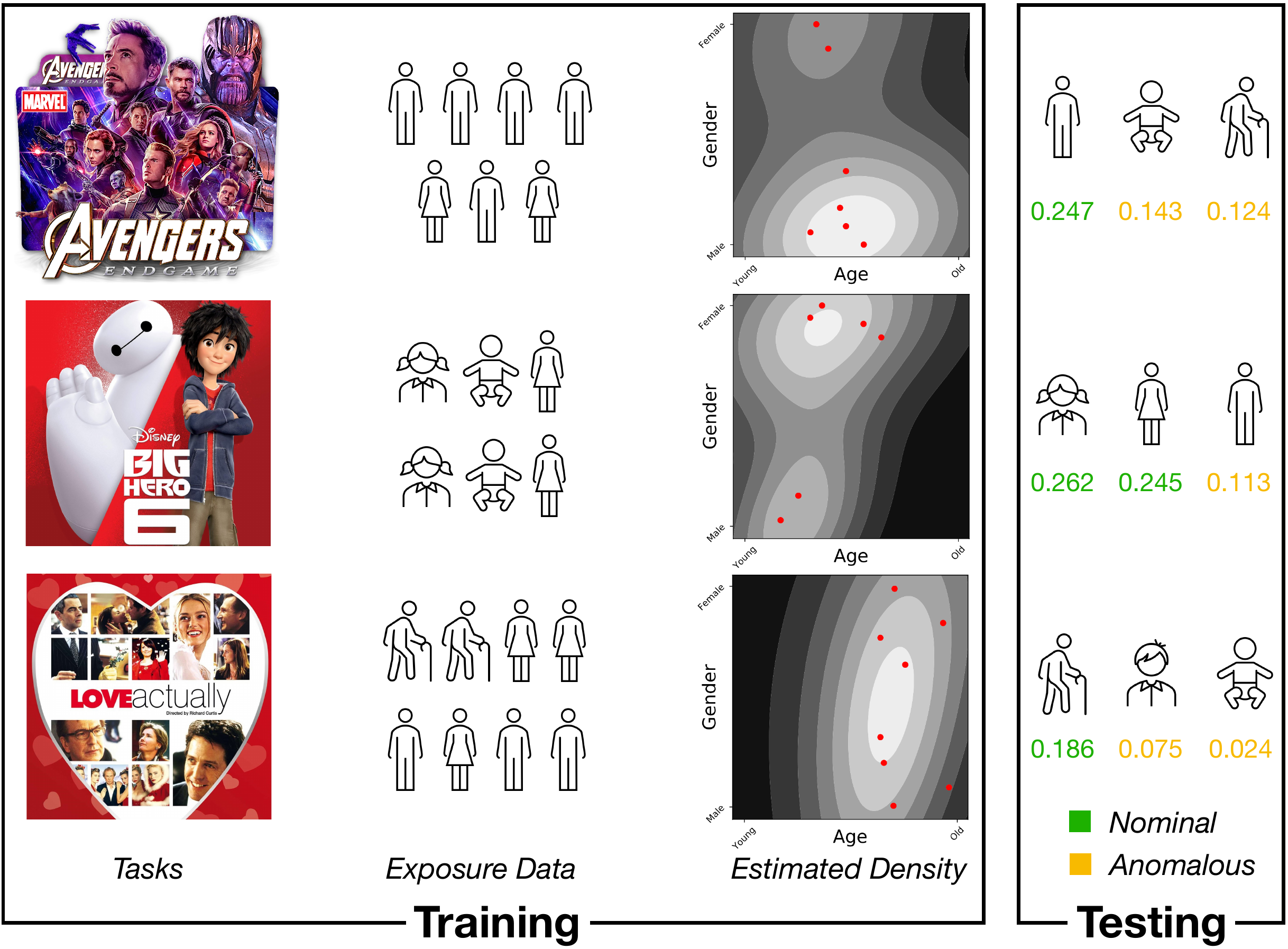}
  \end{minipage}
  }\hfill
  \begin{minipage}[c]{0.48\textwidth}
    \caption{\
       Each movie (task) has audiences with two real-valued features: age and gender.
       The goal is to learn a function for each movie to distinguish between nominal and anomalous audiences in the test set. The third column is an illustration of density estimation over the second column exposure data given the movie. The number under each cartoon portrait on the fourth column is its estimated density, describing the familiarity of this user to the movie. 
    %   We show the density estimation method where the distribution of the audience is estimated given the movie during the training. 
    %   In testing, the probability of a new user is used to identify anomaly. 
    %   It is inefficient to learn a conditional density function for each task restricting it to its exposure data.
    %   Information sharing across tasks has the potential to alleviate such inefficiency.
       %Collaboration among tasks is needed
    }\label{fig:showcase}
  \end{minipage}
 
    \vspace{-10pt}

\end{figure}

We formulate this familiarity estimation problem as a collection of correlated anomaly detection (AD) tasks.
% We illustrate our setting in
Figure~\ref{fig:showcase} illustrates the setting using movie recommendation as an example.
% Assume each movie has been exposed to a group of audiences with two features: age and gender
% The goal is to distinguish for each movie, which users in the test set are ``common'' (nominal), and which are ``fresh'' (anomalous).
Assume that each movie is exposed to audiences with two features: age and gender;
our goal is to determine which users in the test set are ``common'' (\ie nominal) and which users are ``fresh'' (\ie anomalous) for each movie.
In row one of Figure~\ref{fig:showcase}, the movie ``Avengers'' is mostly exposed to young adults, hence preschoolers and the elderly are anomalous to it.
Alternatively, ``Big Hero 6'' is commonly exposed to young audiences, so preschoolers are nominal to it.

Note that the ``age'' and ``gender'' features are for illustration  and evaluation only. These privacy data may not be available in reality. What's more, the items features are usually more complex, {\em e.g.} image or text feature extracted from the pre-trained neural network. Simple statistics like counting over categories are not applicable. 
% Like the conventional anomaly detection problem, we use these tags in evaluation to check the performance of AD.

If each movie is treated as a single AD task, then different movies (tasks) have user groups with different exposure patterns.
Given that tasks share the same user pool, we will consider all tasks collectively instead of individually.

%Note that the ``age'' and ``gender'' label here are only to differentiate the anomaly and nominal data. Like the conventional anomaly detection problem, the category is required in the evaluation  phase.
%What's more, these low-dimension categories are not always available, the item could be a picture, a text description or a high-dimension hash tag. And the features are extracted from the large pre-trained model. Simple statistics like counting on the feature is not applicable.

%If each movie is treated as a single AD task, then different movies (tasks) have user groups with different exposure patterns.
%Given that tasks share the same user pool, we will consider all tasks collectively instead of individually.

There is a plethora of research on the single-task AD problem.
One can estimate the density \(q(\xv)\) of nominal data or a proxy for it~\citep{polonik1995measuring,tsybakov1997nonparametric}, and then set a threshold for the density function to discriminate nominal and anomalous data.
One can also learn a mapping of nominal data from a high-dimensional space to a compact region concentrated around a pre-defined centroid, using kernel functions~\citep{scholkopf2001estimating,tax2004support} or deep neural networks~\citep{ruff2018deep}.
Moreover, outlier exposures are informative to classify nominal {\it vs.} anomalous data~\citep{hendrycks2018deep}.
There are also semi-supervised~\citep{ruff2019deep}, self-supervised~\citep{hendrycks2019using} and fully supervised~\citep{ruff2020rethinking} approaches.
%  demonstrate significant improvement over unsupervised counterparts.

Yet none of the aforementioned methods are tractable to solve all tasks at once when the number of tasks is large.
For example, density-based AD methods seek to estimate density functions for each movie as shown in the third column of Figure~\ref{fig:showcase}.
However, it is inefficient to learn these conditional densities independently, due to the scarcity of exposure data for each task and the linear growth in computation costs with the number of tasks.

We propose Collaborative Anomaly Detection (CAD) to address all tasks simultaneously, where an important observation is that having a massive number of tasks facilitates the training of single tasks.
In CAD, all tasks first learn their initial \emph{task embeddings} to encode the task correlations.
% Then we explore supervised learning via conditional likelihood ratio estimation, which discriminates between exposure data of the current item (positive samples), and exposure data from the other tasks (negative samples).
Then, to get the anomaly detection result, we apply the conditional likelihood ratio estimation for all tasks, where the ratio is between the distribution of exposure data from the current task and the distribution of data from all tasks.
% The samples from the whole distribution further encourage ``collaboration'' between tasks.
We compared our method with baselines such as conditional density estimation with Gaussian models or normalizing flows~\citep{rezende2015variational}.
Our study reveals: \(i\)) likelihood ratio estimation achieves better AD results since it benefits from the full usage of all available data, while being more efficient than density estimation when data are scarce; and
\(ii\)) task embedding initialization is important to effectively exploit task correlations, and we utilize a warm-start strategy with the initial task embedding learned from a small subset of tasks.
% It is beneficial to select a small number of tasks in advance to learn a task embedding model, and then use it to  all the following tasks.
With these properties, CAD scales better and %can
easily adapt to new tasks with limited exposure data. Our problem is motivated from recommendation system, the multi-task anomaly detection setting and the proposed methods can be applied to any other domains.

% reveals two key components of achieve CAD
% First, one need to explicitly borrow data across tasks
% As a first attempt, we try to estimate the density functions conditioning on a ``task embedding'' using either Gaussian mixture models or more flexible models such as normalizing flows~\citep{rezende2015variational}. Our study finds that 1) Using likelihood ratio as a proxy of density enjoys more efficient learning and better results. 2) It is beneficial to select a small number of tasks in prior and encode each task embedding on the coordinate of those ``seed tasks''. With those improvements, CAD scales better and can quickly adapt to new tasks with only a few exposure data.

The remainder of the paper is organized as follows.
Section~\ref{sec:related} briefly reviews related work on AD\@.
Section~\ref{sec:cad} introduces CAD with density estimation and likelihood ratio estimation conditioning on task embeddings.
In Section~\ref{sec:task_embeddings} we discuss how to initialize task embeddings at low cost, and present the entire training procedure.
Empirical results on MovieLens 1M in Section~\ref{sec:exp} show the large efficiency and accuracy gains of CAD relative to naively applying AD for each task.
We also create synthesized tasks on CIFAR10, and test different CAD methods to further study their efficiency and generalization. 
% We further illustrate how to utilize adapt CAD into cold-start recommendation problem. 
Concluding remarks are presented in Section~\ref{sec:conclusion}.

\section{Related Work}\label{sec:related}
Modern AD research has seen significant progress with the utilization of deep learning, where previously common feature mapping implementations such as kernel methods~\citep{scholkopf2001estimating,tax2004support} are replaced by deep networks~\citep{ruff2018deep}.
Deep AD methods either aim to estimate the density of nominal data~\citep{an2015variational,chen2017outlier,huang2019inverse, zisselman2020deep}, or to classify between nominal and anomalous data in semi-supervised, self-supervised or supervised settings~\citep{hendrycks2018deep,hendrycks2019using,ruff2019deep, ruff2020rethinking}.

CAD is a special case of conditional AD, where anomalous data are studied in a context such as time~\citep{gupta2013outlier}, space~\citep{schubert2014local}, or graph structures~\citep{akoglu2015graph}.
CAD studies anomalous data in the context of each item in recommendation systems.
The large number of items and complex latent structures among them make CAD considerably more challenging.

\section{Collaborative Anomaly Detection (CAD)}\label{sec:cad}
\begin{figure*}
\scalebox{0.9}{
\begin{minipage}[t]{0.5\textwidth}
    \vspace{10pt}
    \begin{algorithm}[H]
        \begin{algorithmic}
            \Function {DataSampler}{$M,N$}
            \State Sample \(N\) task indices \(t_i\propto m^{(i)}\), where \(i\in[M]\).
            \State Sample \(N\) instances \(\xv_i\) given \(t_i\) as \(\xv_i\sim q^{(t_i)}(\xv)\).
            \State Sample \(N\) contrastive instances \(\tilde{\xv}_i\sim p(\xv)\).\\
            \quad \ \ \Return \(\mathbf{t}=(t_{1:N}), \xv=(\xv_{1:N}), \tilde{\xv}=(\tilde{\xv}_{1:N}).
            \)
            \EndFunction

            \State
            \Function {EstimateCLR}{$M,\ev$}
            \State Randomly initialize \(\thetav\).
            \State Randomly initialize \(\ev\!\!=\!\!(\!\ev^{(1)},\ldots,\ev^{(M)}\!)\) if \(\ev\!=\!\textproc{None}\).

            \While {not converge}
            \State \(\mathbf{t}, \xv, \tilde{\xv}\) = \Call{DataSampler}{$M$, $N = \textit{batch-size}$}
            \State Update parameters with gradients \(\nabla_{(\thetav,\ev)} L\), where
            \begin{align*}
                L = \frac{1}{N}\sum_{i\in [N]}\Big( & \log(1+\exp(-f(\xv_i,\ev^{(t_i)},\thetav)))              \\
                +                                   & \log(1+\exp(f(\tilde{\xv}_i,\ev^{(t_i)},\thetav)))\Big).
            \end{align*}
            \EndWhile
            \State \Return \(\thetav,\ev\).
            \EndFunction
        \end{algorithmic}
        \caption{Conditional likelihood ratio estimation.}\label{alg:mtocc}
    \end{algorithm}
\end{minipage}\hspace{40pt}
}
\begin{minipage}[t]{0.4\textwidth}
    \vspace{1pt}
    % \hspace{1pt}
    \centering
    \includegraphics[width=0.7\textwidth]{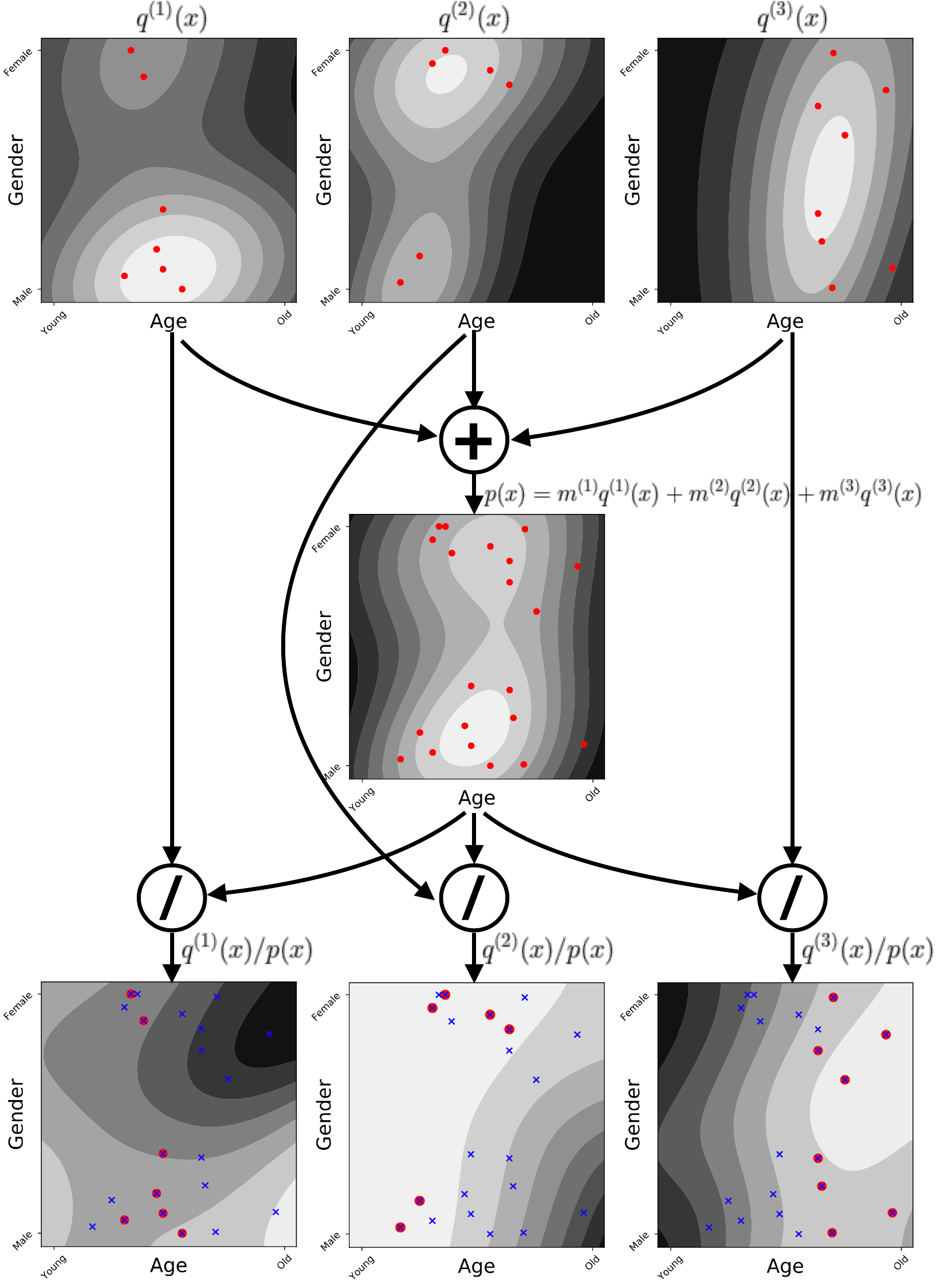}\vspace{-8pt}
    % \caption{(Top) Conditional density. (Middle) Population density. (Bottom) Likelihood ratio.}\label{img:ll}
    % \vfill
  \begin{minipage}[t]{1\textwidth}
    \caption{Colored dots are the samples. Each sample is represented by $1$-d age and gender embedding. (Top) Conditional density of three tasks, (Middle) Base distribution: population density. (Bottom) Likelihood ratio. red dots illustrate the positive sample and blue dots means negative }\label{fig:ll}
  \end{minipage}
\end{minipage}
\end{figure*}
In CAD, each item \(t\) forms a single task, with samples \(\vx\) from its own nominal distribution \(q^{(t)}(\xv)\) representing its exposed users.
Before proceeding to our main approach for CAD using conditional likelihood ratio estimation, we first present and analyze an alternative approach based on conditional density estimation, which will serve as a  baseline.
% We consider two different approaches for CAD, namely, one with conditional density estimation, the other with conditional likelihood ratio estimation.
% We treat the first approach as a strong baseline for the second one.

% Figure \ref{img:ll} demonstrates conditional density functions \(q^{(t)}(x)\) of three tasks, where each user x contains two real-valued 1-d embeddings representing its age and gender separately. To learn those density functions using very few samples (illustrated with small red dots in each plot in the first column) is arguably difficult given the scarcity of data. Instead, one can estimate conditional likelihood ratios with discriminative learning, where each task has its own positive samples (illustrated with red dots in each plot in the third column) and some negative samples (illustrated with blue dots in each plot in the third column) drawn from the population density function (the plot in the second row). The likelihood ratio is learned by contrasting those positive and negative samples. The data augmentation scheme (involving negative samples) makes the likelihood ratio easier to learn.

% One can also observe that density functions have more modes than the likelihood ratio, thus harder to learn. This is because the population density p(x) in this case can smooth out modes in \(q^{(t)}(x)\) and result in simpler and smoother likelihood ratios

\subsection{CAD with conditional density estimation}

In this approach, the goal is to learn a conditional density function \(q(\xv|\ev^{(t)})\) to approximate \(q^{(t)}(\xv)\), and then identify anomalous \(\mathcal{A}^{(t)}\) as data whose density is below a threshold \(\mathcal{A}^{(t)}=\{\vx\in\mathcal{X}:q(\xv|\ev^{(t)})\leq \tau\}\).
Here \(\ev^{(t)}\) is a \emph{task embedding} vector for task \(t\).
Existing methods for conditional density estimation include Gaussian mixture models (GMM), kernel density estimators (KDE),
and those based on deep neural networks such as variants of variational auto-encoders (VAE)~\citep{pol2019anomaly}, generative adversarial networks (GAN)~\citep{mirza2014conditional}, and normalizing flows~\citep{abdelhamed2019noise}.
% To illustrate the procedure, b
Section~\ref{sec:exp} will present experiment results on a variety of those methods.
Below we detail the normalizing flow, as it maximizes the exact likelihood function.

Conditional normalizing flows learn task-specific invertible mappings \(f_{\ev^{(t)}}(\zv):\mathcal{Z}\rightarrow\mathcal{X}\) from a simple task-dependent Gaussian distribution \(\pi_{\ev^{(t)}}(\zv)\) to the target distribution \(q(\xv|\ev^{(t)})\).
The loss function is the negative log-likelihood
\begin{align*}
        L & = -\log q(\xv|\ev^{(t)}) \\
        & = -\log\pi_{\ev^{(t)}}(f^{-1}_{\ev^{(t)}}(\xv)) - \log\left|\det\frac{\partial f^{-1}_{\ev^{(t)}}(\xv)}{\partial\xv}\right|
\end{align*}
% .
% \]
In practice, flexible normalizing flows are implemented by stacking invertible layers \(f_{\ev^{(t)}}(\zv)=f^K_{\ev^{(t)}}\circ f^{K-1}_{\ev^{(t)}}\circ\ldots\circ f^1_{\ev^{(t)}}(\zv)\)~\citep{winkler2019learning,abdal2021styleflow}.
Our experiments suggest that conditional normalizing flows suffer from two limitations.
\(i\)) Estimating \(q^{(t)}(\xv)\) is hard since \(\xv\) lies in a high-dimensional space and there are often only a few samples available for each task.
Further, the estimated conditional density function may assign a high score for anomalous data due to the inductive bias of normalizing flows~\citep{nalisnick2018deep,kirichenko2020normalizing}.
% $ii$) The performance of density estimation highly depends on the initialization of $\ev^{(t)}$.
\(ii\)) The performance of density estimation highly depends on the number of tasks.

To alleviate these issues, in Section~\ref{sec:cad_alg} we consider estimating the conditional likelihood-ratio. Then in Section~\ref{sec:task_embeddings} we propose to use seed tasks to initialize the task embeddings.

\subsection{CAD with conditional likelihood ratio estimation}\label{sec:cad_alg}

We consider introducing a \textit{base distribution} \(p(\xv)\) from which we can sample.
Then, CAD estimates the log-likelihood ratio \(r^{(t)}(\xv)\) as a proxy for the density \(q^{(t)}(\xv)\) 
\begin{align}\vspace{-3pt}
    r^{(t)}(\xv)=\log \frac{q^{(t)}(\xv)}{p(\xv)}.\vspace{-3pt}
\end{align}

\paragraph{Base distribution selection}
As illustrated in Algorithm~\ref{alg:mtocc}, the base distribution resembles the negative samples in the one-class classification literature~\citep{steinwart2005classification}.
A good choice for the base distribution varies with the application; however, it should be informative enough to characterize the boundary between nominal and anomalous data.
In the context of recommendation systems, we choose the base distribution as the population exposure distribution \(p(\xv)=\sum_t m^{(t)}q^{(t)}(\xv)\), where \(m^{(t)}\) is the proportion of users exposed to task \(t\).
Let \(\mathcal{P}\) be a feasible domain of probability measures.
We can prove that
\begin{proposition}\label{prop:kl}
    \(p(\xv)=\sum_t\!m^{(t)}q^{(t)}(\xv)\) is optimal for minimizing the expected KL divergence
    % \begin{small}
    % \begin{align}\label{eq:min_kl}
    %     \min_{p(\xv)\in\mathcal{P}}\mathbb{E}[\operatorname{KL}(q^{(t)}|p)] =\min_{p(\xv)\in\mathcal{P}}\sum_t\mathbb{E}_{q^{(t)}(\xv)} \left[ m^{(t)} \log \frac{q^{(t)}(\xv)}{p(\xv)} \right].\vspace{-3pt}
    % \end{align}
    % \end{small}
    \resizebox{.95\linewidth}{!}{
  \begin{minipage}{\linewidth}
  \begin{align}
\min_{p(\xv)\in\mathcal{P}}\mathbb{E}[\operatorname{KL}(q^{(t)}|p)] =\min_{p(\xv)\in\mathcal{P}}\sum_t\mathbb{E}_{q^{(t)}(\xv)} \left[ m^{(t)} \log \frac{q^{(t)}(\xv)}{p(\xv)} \right]
\end{align}
  \end{minipage}
}
\end{proposition}

% \begin{proof}
% See the Appendix.
% \end{proof}

Instead of using the standard density estimation defined on the Lebesgue measure \(\mathcal{R}^n\), our likelihood ratio is a Radon–Nikodym derivative \wrt \(p(\xv)\), which has two benefits.
First, the likelihood ratio corrects the bias of the density by the population.
For example, a moderately common sample \(\vx\) in one task \(t\), namely \(q^{(t)}(\xv)\approx 1\), should be marked as significant even when it is extremely rare in the population, \ie, \(p(\xv)\approx 0\).
Second, previous work on computational stability~\citep{friedman2001elements} shows that a smaller distance between two distributions leads to more accurate likelihood-ratio estimation.
Moreover, Proposition~\ref{prop:kl} shows that our choice of \(p(\xv)\) is the best option under the Kullback–Leibler (KL) divergence metric. The proof can be found in the appendix.

Figure \ref{fig:ll} demonstrates conditional density functions \(q^{(t)}(x)\) of three tasks. To learn the density functions is arguably difficult given the scarcity of data (first row). Instead, one can estimate conditional likelihood ratios with discriminative learning, where each task has its own positive samples and some negative samples (third row) drawn from the population density function (second row). 
% The data augmentation scheme (involving negative samples) makes the likelihood ratio easier to learn. 
One can also observe that density functions have more modes than the likelihood ratio, thus harder to learn. This is because the population density p(x) in this case can smooth out modes in \(q^{(t)}(x)\) and result in simpler and smoother likelihood ratios.

\paragraph{Likelihood ratio estimation}

% We learn the ratio \(r^{(t)}(\xv)\) with supervised methods.
We learn the ratio \(r^{(t)}(\xv)\) by logistic regression.
Estimating the likelihood ratio for a single task \(t\) is considered first, and then we extend this to jointly learn all tasks.
For each task \(t\), we sample an observation from the nominal distribution \(\xv\sim q^{(t)}(\xv)\) as a positive example and pair it with a negative example drawn as \(\tilde{\xv}\sim p(\tilde{\xv})\).
Repeatedly doing so, we collect \(N\) contrastive pairs \((\xv_i,y_i=1)\) and \((\tilde{\xv}_i,\tilde{y}_i=-1)\).

When \(N\) goes to infinity, the optimal solution \(\thetav^*\) for the logistic regression loss
\begin{align}
    \label{eq:embed}
    L(\thetav) = & \frac{1}{N}\Big(\sum_{i\in [N]}\log(1+\exp(-f(\xv_i,\thetav)))\nonumber \\
    &+ \sum_{i\in [N]}\log(1+\exp(f(\tilde{\xv}_i,\thetav)))\Big),
\end{align}
guarantees that \(f(\xv,\thetav^*)=\log (q^{(t)}(\xv)/p(\xv))\) for any \(\xv\in\operatorname{supp}(q^{(t)})\)~\citep{huang2006correcting}. The proof is in the appendix.

To learn all tasks at once, assume the number of users exposed to item \(t\) is proportional to \(m^{(t)}\) and \(\sum_t m^{(t)}=1\).
We sample a random user \(\xv\) by first sampling her/his task assignment \(t\propto m^{(t)}\) and then sampling \(\xv\sim q^{(t)}(\xv)\).
The construction of contrastive pairs naturally extends from the single-task case.
In our implementation, \(f(\xv,\ev^{(t)},\thetav)\) is a deep neural network parameterized by \(\thetav\), and each task is characterized by a vector \(\ev^{(t)}\), \ie the task embedding.
We demonstrate our learning framework in Algorithm~\ref{alg:mtocc} as \(\textproc{EstimateCLR}(M,\ev)\), where \(M\) is the number of tasks and \(\ev\) stands for a task embedding initialization.
We use random initialization when setting \(\ev=\textproc{None}\), which is typically not the best choice in practice.
We will discuss better initialization strategies in Section~\ref{sec:task_embeddings}.

% Anomaly detection aims to identify out-of-distribution data that deviates from a nominal, target distribution \(q(\xv)\).
% In many scenarios, AD needs to be customized for each individual user or group, resulting in a massive set of correlated tasks.
% For example, in recommendation systems, each item needs to maximize its own utility by {\it exploiting\/} its existing user pool (the nominal distribution) and {\it exploring\/} potentially new users (out-of-distribution data).
% Each item \(t\) forms a single task, with its own nominal distribution \(q^{(t)}(\xv)\) representing the distribution of its users.
% Estimating \(q^{(t)}(\xv)\) is particularly hard since \(\xv\) usually lives in a high-dimensional space and there are very few samples available for each task.

% Collaborative Anomaly Detection (CAD) considers an alternative to the density function by introducing a base distribution \(p(\xv)\) from which we can sample.
% Then, CAD estimates the log-likelihood ratio
% \[
%     r^{(t)}(\xv)=\log \frac{q^{(t)}(\xv)}{p(\xv)} ,
% \]
% as a proxy for the density $q^{(t)}$.

\section{Task Embeddings Initialization}\label{sec:task_embeddings}

In this section, we focus on \textit{learning} a tailored task embedding \(\widehat{\ev}^{(t)}\) that can be used in initialization of Algorithm~\ref{alg:mtocc}.
We seek task embeddings with the following properties:
\begin{enumerate}
    \item (\textbf{Correlation}) 
    % Similar to, we find that randomly initialized task embeddings do not fully exploit task similarities during training, even with fine-tuning.
    % To overcome this, t
    The learned initialization should capture task similarity.~\citep{meyerson2020traveling} 
    \item (\textbf{Generalization}) The learned task embedding models can generalize to new correlated tasks.
    \item (\textbf{Efficiency})
    % In CAD, each task \(t\) is identified as a distribution \(q^{(t)}(\xv)\), which can well capture the correlation bewteen tasks. 
    % A natural measurement of task similarity is to calculate their distributional distances.
    % However, this requires an enormous number of samples to approximate all pair-wise distances.
    % By contrast,
    The acquisition of learned task embeddings should be inexpensive.
\end{enumerate}

% \subsection{Training procedure}

Suppose there are \(M\) tasks in total.
We propose to learn the task embedding \(\widehat{\ev}^{(t)}\) by first selecting \(M_0\ll M\) \emph{seed tasks} to train a \emph{pre-embedding model}, and then use the pre-embedding model to encode all tasks.
In practice, we find that randomly selected seed tasks are representative enough to capture correlations among all tasks.
This is hypothetically due to the redundancy among tasks where the inherent complexity of semantics may not be that big.
We study the choice of \(M_0\) in Section~\ref{sec:mnist_cifar}.

\paragraph{Train pre-embedding model with seed tasks}
Without loss of generality, seed tasks are identified as task \(1\) to \(M_0\).
We train a log-likelihood ratio task to get the pre-embedding model by calling \(\psi,\ev=\textproc{EstimateCLR}(M_0,\textproc{None})\) in Algorithm~\ref{alg:mtocc}, where \(\psi\) denotes the parameters of the pre-embedding model (not related to the final conditional likelihood ratio estimation model).
The returned vector \(\ev\) is dropped,
and the pre-embedding model \(\psi\) is sent to the next step.

\paragraph{Obtaining task embeddings initializations}
Given the pre-embedding model \(\psi\), the following basis functions are constructed,
\begin{equation}\label{eq:response_vec}
    \vr(\vx)=\left(\vr^{(1)}(\vx), \ldots, \vr^{(M_0)}(\vx)\right),\nonumber
\end{equation}
where \(\vr^{(s)}(\xv)=\log(q^{(s)}(\xv)/p(\xv))\) is the log-likelihood ratio of the pre-embedding model \(\psi\).
The task embedding \(\widehat{\ev}^{(t)}\) is defined as a linear projection of \(q^{(t)}(\xv)\) onto these basis functions, \ie,
\begin{align}\label{eq:moh_embeeding_vec}
\widehat{\ev}^{(t)} = \left(\mathbb{E}_{q^{(t)}(\xv)}[\vr^{(1)}(\xv)],\ldots, \mathbb{E}_{q^{(t)}(\xv)}[\vr^{(M_0)}(\xv)]\right).
\end{align}
We independently draw \(N^{(t)}\) samples \(\xv_1,\ldots,\xv_{N^{(t)}}\) from \(q^{(t)}(\xv)\) to estimate \(\widehat{\ev}^{(t)}\), where each entry is estimated by \(\mathbb{E}_{q^{(t)}(\xv)}[\vr^{(s)}(\xv)]\approx\frac{1}{N^{(t)}}\sum_{n=1}^{N^{(t)}} f(\xv_n,\ev^{(s)},\psi)\).
This step only requires forward passing and does not involve any model updates.

After obtaining the learned task embedding initializations \(\widehat{\ev}=(\widehat{\ev}^{(1)},\ldots,\widehat{\ev}^{(M)})\), we can run the conditional likelihood ratio estimation as described in Section~\ref{sec:cad_alg}, where
 \(\theta,\ev=\textproc{EstimateCLR}(M,\widehat{\ev})\).
This completes the algorithm.

\subsection{Interpret the learned embeddings}

To informally explain why the task embedding initialization \(\widehat{\ev}^{(t)}\) in~(\ref{eq:moh_embeeding_vec}) captures task correlations, we treat \(q^{(t)}(\xv)\) and \(r^{(s)}(\xv)\) as infinite-dimensional vectors by varying \(\xv\) across its support.
% , we collect $\{q^{(t)}(\xv)\}_{t\in [M]}\subset\mathcal{P}$ with their spanned space $\mathcal{Q}$.
% The dual space of $\mathcal{Q}$ consists of integrable functions on $\mathcal{Q}$.
% Intuitively, it is preferable to work with the dual space since the space of integrable functions is more flexible to explore than the original probability distribution space.
% When treating $q^{(t)}(\xv)$ and $r^{(s)}(\xv)$ as infinite-length vectors by varying $\xv$ across its support,
Random-projection theory guarantees that \(\text{sim}(\widehat{\ev}^{(t)},\widehat{\ev}^{(t')})\approx\text{sim}(q^{(t)},q^{(t')})\), when \(M_0\rightarrow\infty\) with random regular basis functions \(\vr(\xv)\) and properly chosen similarity metrics~\citep{gottlieb2015nonlinear}.
In practice, naively generated random functions may not work well, since the random function must capture the difference of \(q^{(t)}\) and \(q^{(t')}\) on their support.
Likelihood ratios are more reasonable to cover all supports, while still being discriminative since they are learned by contrastive pairs on the population support \(\text{supp}(p(\xv))\).
In our experiments, we found the number of basis functions \(M_0\) can be much smaller than the total number of tasks \(M\) when tasks are correlated and redundant.
As we will show in experiments, the learned embedding can also generalize to previously unseen tasks.

% When user attributes or side information are available, it is possible to improve task embedding quality with those attributes.
% We defer this discussion to the experiments and highlight that our method can learn task embeddings in the absence of attributes.

% \input{coldstart}
\section{Experiments}\label{sec:exp}

We start with experiments on the MovieLens 1M dataset to demonstrate that \(i\)) CAD with conditional likelihood ratio estimation achieves much better results compared to conditional density estimation; and \(ii\)) task embedding initialization is crucial.
To further analyze these properties, we design a set of experiments on an image dataset (CIFAR10) with synthesized tasks.
Unless otherwise stated, all experiments are executed on a single NVIDIA Titan Xp GPU with 12,196M Memory.
% image datasets by creating tasks with different image label preferences.

\subsection{Experiments on MovieLens 1M}\label{subsec:movielens}

\paragraph{Problem setup}
The MovieLens 1M dataset contains approximately \(3,900\) movies and \(6,040\) users, together with over one million ratings~\citep{harper2015movielens},
hence \(3,900\) AD tasks in total, one for each movie.
We ignore all rating scores and only keep the binary information of whether movies are rated (exposed) or not.
We split users into training and testing sets with 8:2 ratio.
The goal for each AD task is to distinguish between ``common users'' (nominal) and ``fresh users'' (anomalous) in the testing set. 
% Note that we can also do anomaly detection on items given the user by replacing the position of user and item. 

\begin{table*}[t]
\caption{Anomaly detection results in terms of AUC\;(\(\times\)100). (Left) Comparing GMM, normalizing flows, and likelihood-ratio estimation. (Right) Conditional likelihood ratio estimation with various task embedding initialization methods. * means extra knowledge is used for initialization.}\label{table:movielens}\vspace{-5pt}
\begin{minipage}{0.49\textwidth}
\begin{center}
\begin{small}
\begin{sc}
\scalebox{0.95}{
\begin{tabular}{l|cc}
\toprule
Method & Age & Occupation \\
\midrule
Gaussian & 59.5 & 61.5 \\
Normalizing Flow & 60.9  & 58.3\\
Likelihood Ratio & \textbf{78.2} & \textbf{64.2} \\
\bottomrule
\end{tabular}
}
\end{sc}
\end{small}
\end{center}
\end{minipage}\hspace{5pt}
\begin{minipage}{0.45\textwidth}
\vspace{5pt}
\begin{center}
\begin{small}
\begin{sc}
\scalebox{0.9}{
\begin{tabular}{l|cc}
\toprule
Embedding Initialization & Age & Occupation \\
\midrule
Graph Embedding* & 80.3 & 68.0 \\
Histogram Embedding* & \textbf{82.6} & \textbf{73.4} \\
% Two label Embedding*  & 81.4          & 69.0          \\
\midrule
Random Initialization & 78.2 & 64.2 \\
Learned Embedding & \textbf{80.6} & \textbf{66.1} \\
\bottomrule
\end{tabular}
}
\end{sc}
\end{small}
\end{center}
\end{minipage}
\end{table*}

\paragraph{Feature pre-processing}
In MovieLens 1M data, there are four categorical labels for users: gender \((u_g)\), age \((u_a)\), occupation \((u_o)\), and zip-code \((u_z)\); and three categorical features for items: movie id, title, and genre.
We use a deep factorization-machine~\citep{guo2017deepfm} to extract user embeddings \(\xv\) (details in appendix) for training AD models.
The detailed four categorical labels \(u_{n,g},\;u_{n,a},\;u_{n,o},\;u_{n,z}\) for user \(\xv_n\) will be used for testing only.

\paragraph{Nominal and anomalous}
The definition of anomalous varies across domains~\citep{ruff2021unifying}.
We consider point anomaly in the context of recommendation systems, which aims to identify each anomalous user for an item \(t\).
An anomalous user is defined according to a specific user label.
Take the label ``age'' as an example.
Suppose an item \(t\) is exposed to a set of users \(\Xv^{(t)}=\{\xv^{(t)}_1,\ldots,\xv^{(t)}_{m^{(t)}}\}\), where the age of user \(\xv^{(t)}_n\) is \(u^{(t)}_{n,a}\). We identify the most frequent age category \(\hat{u}^{t}_a:=\arg\max_u \sum_{n=1}^{m^{(t)}}\mathbf{1}(u_{n,a}^{(t)}=u)\) as nominal and the least frequent age category \(\check{u}^{t}_a:=\arg\min_u \sum_{n=1}^{m^{(t)}}\mathbf{1}(u_{n,a}^{(t)}=u)\) as anomalous.
The AUC score is calculated on the test set by distinguishing between the nominal and anomalous users only.
Since there are multiple user labels, we may report multiple AUC scores, one for each user label.
In our experiments below, we test on two user labels: ``Age'' with 5 binned categories and ``Occupation'' with 21 categories.

\paragraph{Task filtering}
We remove tasks that are not eligible for AD experiments.
To be precise, we first remove tasks with fewer than 100 exposures.
Among the remaining tasks, we select \(50\%\) of them with the lowest user label histogram entropy, indicating nominal and anomalous are distinguishable.
There are 1,629 remaining tasks after the filtering.

% \footnote{We use the implementation in \hyperlink{https://github.com/TaoMiner/joint-kg-recommender}{https://github.com/TaoMiner/joint-kg-recommender}}.
% feature matching with user-movie connections and movie knowledge graphs~\citep{cao2019unifying}.

\paragraph{Methods}
We study CAD with the proposed conditional likelihood ratio estimation method and two conditional density estimation methods, conditional Gaussian and conditional normalizing flows.
For conditional likelihood ratio estimation, we implement a 3-layer MLP, where task embeddings are concatenated to the input.
For Gaussian, we use a 3-layer MLP with two output heads to map the task embedding to Gaussian mean and variance.
For normalizing flows, we follow~\citep{papamakarios2017masked} to stack \(5\) blocks,
where each block is a stack of MADE~\citep{germain2015made} and BatchNorm.
For all methods, we train until their losses do not decrease, and select the best checkpoints on validation and report results on testing data.
Implementation details are put in %\textcolor{red}{Appendix}.
appendix.

The AUC results are shown in Table~\ref{table:movielens} (left).
Conditional density estimation completely fails in the CAD scenario, on both ``Age'' labels and on ``Occupation'' labels.
Normalizing flows are much more flexible than Gaussian, yet still not able to discriminate the anomalous.
This is because density estimation usually requires large data and model capacity for learning.
When conditioning on too many tasks, density estimation may fail to distinguish the complex correlations among tasks.
This phenomenon will be further inspected in Section~\ref{sec:mnist_cifar}.

% Put this in appendix.
% The MT-OCC network uses user features and task (movie) embeddings as network inputs, followed by a three-layer fully connected network with ReLU activation functions.

\paragraph{Task embedding initialization}
In conditional likelihood ratio estimation, task embedding \(\ev^{(t)}\) plays a critical role for data sharing across tasks, since learning one task can enhance the learning of another related task with a similar task embedding.
We compare the following choices of task embedding initialization.
%
%\begin{enumerate}[label=(\alph*)]
\(i\)) \textit{Graph Embedding}: use learned movie features through a knowledge graph base using KTUP~\citep{cao2019unifying} to initialize \(\ev^{(t)}\).
\(ii\)) \textit{Histogram Embedding}: when training on age labels, we calculate the histogram of users' age labels for each task \(t\): \(\{u^{(t)}_{1,a},\ldots,u^{(t)}_{m^{(t)},a}\}\) to initialize \(\ev^{(t)}\).
This is similarly done for occupations.
% \item Two-Label Embedding. Concatenate the ``age'' label embedding and ``occupation'' label embedding as \(\ev^{(t)}\).
\(iii\)) \textit{Random Initialization}: use normal distribution to randomly initialize \(\ev^{(t)}\).
\(iv\)) \textit{Learned Embedding}: initialize \(\ev^{(t)}\) following the procedure in Section~\ref{sec:task_embeddings} with \(M_0=100\) seed tasks. We use the same training scheme for all initialization methods. The training detail can be found in the %\textcolor{red}{Appendix}.
appendix.
% Use the multiple output head network to learn the response on a subset of data. The dual embedding is the empirical average over responses from task $t$.
% \end{enumerate}

% \textcolor{red}{learning rate, training epochs.}

% Put this into appendix.
% We build a three-layer fully connected neural network with ReLU activation function.
% For each method, the initial learning rate is \(1 \! \times \! 10^{-4}\), which further decreases by 0.1 at epoch 80 and 100 consecutively.
% We train 120 epochs in total and select the model with the highest AUC on the validation dataset as our final model.
% The embedding size for random initialization and explicit feature is 16.
% The dimension of label embeddings depends on the number of categories used.
% For the dual embedding, we randomly choose $M_0 = 100$ task responses as the task representation.
% The label and task embeddings have an extra linear mapping between the embedding size to 16 for consistency with the randomly initialized embedding.

As shown in Table~\ref{table:movielens} (right), task embeddings with additional knowledge (marked with \(*\)) are generally superior over embeddings without such knowledge.
Learned embeddings are significantly better than random embeddings, and even comparable with graph embeddings, since they attempt to uncover the connections underlying over \(1,600\) tasks using seed tasks.

\paragraph{Downstream tasks}
 So far we have shown that our likelihood ratio can tell the anomaly data from the nominal one. It is a good measure of the familiarity between users and items. We could use this familiarity to diversify the recommendation (explore) or fit the user's current preference (exploit). After collecting feedback like users' reactions, satisfaction rate, or advertisements' revenue, we can re-estimate familiarity and do the next round of recommendations.
Even though this pipeline is nearly impossible to simulate given the current public dataset and evaluation metrics, we do believe that the familiarity between user and item is a necessary and meaningful signal to reflect the user behavior or item property. For more scenarios to use CAD, please check the Appendix.

\begin{table*}[t]
    % \tiny
    \caption{Testing AUC\;(\(\times\)100) on CIFAR10 with increasing difficulty levels $k=1,2,3,4$. The number in the brackets represents the total number of tasks. (Left) Separate learning. (Right) Conditional learning. $^\dagger$ means use pre-trained feature. }\vspace{-7pt}\label{table:cifar10}
    \vspace{6pt}
    \begin{minipage}{0.59\textwidth}
        \begin{center}
            \begin{small}
                \begin{sc}
                    \scalebox{0.73}{
                        \begin{tabular}{c|ccccc}
                            \toprule
                            Method~\textbackslash{}~k & 1(10)                & 2(45)               & 3(120)              & 4(210)               & \#params \\
                            \midrule
                            Gaussian$^\dagger$                  & $93.92 \!\pm\! 3.2$  & $66.45 \!\pm\! 5.4$ & $60.90 \!\pm\! 5.8$ & $54.60 \!\pm\! 6.0$ & 1M                \\
                            % 5 blocks, lr 4e-3, 3 blocks lr 1e-2
                            MAF$^\dagger$                        & $91.96  \!\pm\! 1.7$ & $90.76 \!\pm\! 2.3$ & $83.77 \!\pm\! 2.7$ & $ 79.40 \!\pm\!1.2 $      &  0.4M         \\
                            LRatio$^\dagger$               &    \textbf{99.58}$\pm$\textbf{0.9}           &  \textbf{98.57}$\pm$\textbf{0.9}   &   \textbf{97.61}$\pm$\textbf{1.1}      &       \textbf{95.67}$\pm$\textbf{1.5}  &  0.1M           \\
                            \midrule
                            %GLOW & $94.02\!\pm\! 3.0$ & $89.93\!\pm\! 2.6$  & $88.28 \!\pm\! 0.7$ & $83.72 \!\pm\! 1.2$ \\
                            Deep SVDD                  &  $64.81 \!\pm\! 6.8$      &     $52.43 \!\pm\! 2.3$      &        $51.43 \!\pm\! 5.3$    &    $50.68 \!\pm\! 4.7$       & 0.5M         \\
                            CSI                       & $94.28 \!\pm\! 3.8$          &    $62.53 \!\pm\! 5.3$             &     $55.78 \!\pm\!5.6$                &     $53.78 \!\pm\! 2.3$       &    11.5M        \\
                            LRatio                   &   \textbf{96.05}$\pm$\textbf{2.3}    &   \textbf{84.26}$\pm$\textbf{5.0}              &    \textbf{74.35}$\pm$\textbf{5.1}      &      \textbf{68.05}$\pm$\textbf{5.6}       &  3M     \\
                            \bottomrule
                        \end{tabular}
                    }
                \end{sc}
            \end{small}
        \end{center}
    \end{minipage}\hspace{10pt}
    \begin{minipage}{0.39\textwidth}
    \vspace{-12pt}
        \begin{center}
            \begin{small}
                \begin{sc}
                    \scalebox{0.72}{
                        \begin{tabular}{c|cccc}
                            \toprule
                            Method~\textbackslash{}~k & 1(10) & 2(45) & 3(120) & 4(210) \\
                            \midrule
                            % GMM &88.23 & 74.27 & 74.34 & 63.76 mnist
                            Gaussian$^\dagger$      & 75.76 & 62.52 & 58.21  & 56.10  \\
                            % MAF & 71.12 & 66.15 & 65.33 & 62.87 \\
                            % MAF  & 87.84 & 71.27 & 67.01 & 64.53 \\
                            MAF$^\dagger$            & 92.04 & 82.25 & 80.54  & 61.56  \\
                            % GLOW & 82.75 & 54.32 & 52.15 & 51.17 \\
                            % LRatio &  & \textcolor{red}{97.02} & \textcolor{red}{92.79} & \textcolor{red}{90.98}\\
                            % LRatio &  & \textbf{96.43} & \textbf{95.06} & \textbf{93.45} \\
                            LRatio$^\dagger$ & \textbf{99.41} & \textbf{98.97} & \textbf{98.25} & \textbf{97.46} \\
                            LRatio & 95.97 & 93.07 & 87.13  & 78.96 \\
                            \bottomrule
                        \end{tabular}
                    }
                \end{sc}
            \end{small}
        \end{center}
    \end{minipage}
    \vspace{-3pt}
    \label{table:toy_result}
\end{table*}

\begin{table*}[t]\vspace{-10pt}
\caption{
    Testing AUC\;(\(\times\)100) on CIFAR10. (Left) Comparing various task embedding initialization methods. Label embedding uses categorical label information (marked with a *) so it has a natural advantage comparing with other methods. The best results without extra knowledge are shown in bold. (Right) To test the generalization performance of CAD, we train on a set of tasks (rows) and test on a different set of tasks (columns).
    %* implies that category information is used during training.
}\label{table:toy_result}
\begin{minipage}{0.49\textwidth}
\begin{center}
\begin{small}
\begin{sc}
\scalebox{0.8}{
    \begin{tabular}{ccccc}
        \toprule
        \multicolumn{1}{c|}{Embedding Init.~\textbackslash{}~k} & 2(45)         & 3(120)        & 4(210)        & 5(252)        \\
        \midrule
        \multicolumn{1}{c|}{Label Emb.* \((L\!=\!10)\)}         & 96.50          & 95.13          & 94.21          & 91.90          \\
        \multicolumn{1}{c|}{Pseudo Emb. \((L\!=\!10)\)}         & 96.00          & 94.13          & 92.21          & 85.90          \\
        \multicolumn{1}{c|}{Random Init. \((L\!=\!10)\)}        & 95.97          & 93.07          & 87.13          & 78.96          \\
        \multicolumn{1}{c|}{Learned Emb. \((M_0\!=\!10)\)}           & 96.20          & 94.61          & 92.11          & 90.30          \\
        \multicolumn{1}{c|}{Learned Emb. \((M_0\!=\!64)\)}           & \textbf{96.43} & \textbf{95.06} & \textbf{93.45} & \textbf{91.40} \\
        \bottomrule
    \end{tabular}
}
\end{sc}
\end{small}
\end{center}
\end{minipage}\hspace{15pt}
\begin{minipage}{0.49\textwidth}
\begin{center}
\begin{small}
\begin{sc}
\scalebox{0.8}{
    \begin{tabular}{c|cccc}
        \toprule
        train~m\textbackslash{}test~n & 2(45) & 3(120) & 4(210) & 5(252) \\
        \midrule
        2~(45)                        & ---   & 89.11  & 79.33  & 73.11  \\
        3~(120)                       & 95.80 & ---    & 86.62  & 80.53  \\
        4~(210)                       & 95.26 & 93.94  & ---    & 86.52  \\
        5~(252)                       & 91.96 & 91.61  & 91.08  & ---    \\
        \bottomrule
    \end{tabular}
}
\end{sc}
\end{small}
\end{center}
\end{minipage}
\vspace{-3pt}
\end{table*}

\vspace{-5pt}
\subsection{Experimental studies on CIFAR10}\label{sec:mnist_cifar}

The difficulty of CAD stems from the large number of correlated tasks.
In this section, we synthesize AD tasks using the CIFAR10 dataset~\citep{Krizhevsky09learningmultiple} to gain further understandings of the two phenomena: \(i\)) why conditional likelihood ratio estimation leads to better CAD results; and \(ii\)) why the learned task embeddings work better than randomly initialized embeddings. To connect with previous recommendation setting, we can treat each task as a ``user'', the images in each task as ``items'' exposed to this user. The union of items in all tasks make up the whole dataset, which is in accordance with the recommendation that all the users share the one item set. 

\paragraph{Problem setup}
Without further specification, the experiments below use the default training testing split for CIFAR10.
We further split out \(5,000\) samples out of \(60,000\) training samples as a validation set.
CIFAR10 has ten ground-truth categories.
We will select a subset of categories, called ``active categories'', to expose to a task.
For evaluation, given the task index, we select all the samples from its active category as the nominal data and treat the remaining data as anomalous.
We report the average AUC for all tasks on the test set.

\paragraph{CAD with increasing complexity}
We artificially create a sequence of experiments indexed by  \(1\leq k\leq 4\) with an increasing number of correlated tasks .
% Each experiment is indexed by an integer.
For each task in one experiment, we select \(k\) active categories out of ten.
% The rest $(10-k)$ categories are never exposed to the task.
We enumerate all possible categorical combinations, resulting in \(C_{10}^k\) tasks in total.
Each image is randomly exposed to only one task that is active for training.
The more tasks one has, the fewer samples each task contains.
We prepare four experiments with increasing complexity \(k=1,2,3,4\) on two scenarios: \(i\)) separately learning, all tasks are learned separately; \(ii\)) conditional learning, all tasks are learned jointly with information sharing.
% and information sharing across tasks is required.

We study CAD with two (conditional) density estimation methods: Gaussian and masked autoregressive flow (MAF,~\citep{papamakarios2017masked}) 
and compare with two modern AD methods: deep support vector data description (Deep SVDD,~\citep{ruff2018deep}) and anomaly detection with contrasting shifted instances (CSI,~\citep{tack2020csi}). Gaussian and MAF are implemented on 128d features extracted from an unsupervised model since they work better with high-level, dense features (\citep{kirichenko2020normalizing}).  
% We compare with Deep SVDD and CSI on raw images since CSI relies on image-level operations such as rotation. We implement our (conditional) likelihood ratio estimation (LRatio) both for extracted features and raw images.

Table~\ref{table:cifar10} (left) shows separately learning results.
% Density estimation models need more capacity to characterize complex data distribution.
There is a sharp performance drop for Gaussian when \(k\) grows since it can only capture one mode of data distribution.
MAF is more flexible than Gaussian, thus performs much better.
The performance of all methods (CSI, SVDD and LRatio) trained with raw images drops quickly when the number of tasks increases since fewer samples are assigned to each task.
% , and distinguishing categories on the raw image space is generally harder.
% SVDD is not suitable for discriminating between any group of active categories with other categories since it has no evidence of anomaly during training.
% CSI does better than SVDD since it uses shifted instances as evidence for anomalous.
Likelihood ratio estimation exploits samples from the base distribution, thus better discriminates nominal and anomalous both for extracted features and raw images.
% However, when active categories include several categories with unknown labels, the anomalous and nominal data are mixed

Table~\ref{table:cifar10} (right) lists the conditional learning results.
% Gaussian distribution has limited capacity, it has severe limitations in adapting to different tasks.
Conditional Gaussian is generally as bad as separately trained Gaussian because of limited capacity.
MAF has enough capacity to learn a single task but suffers from large $k$. 
% because conditional MAF relies on information sharing only through task embeddings and parameter sharing, which is implicit.
Conditional likelihood ratio estimation performs well with the help of good task embedding and information sharing from negative samples.
% shares information by borrowing negative samples across tasks as well as using task embeddings,
% which helps it accurately learn each task even when the number of tasks grows.

% \input{figures/cifar120}

\paragraph{Task embedding initialization}
% We take a closer look at different task embedding initialization methods for conditional likelihood ratio estimation,
% To demonstrating that the learned task embeddings can preserved task correlations.
% The detailed architecture used to integrate task embeddings can be found in the \textcolor{red}{Appendix}.
We compare learned task embeddings with the following task embedding initialization approaches.
% Since lack of data for each task, independently train on those tasks are especially inaccurate. We compare several MT-OCC methods using the same architecture with different task embeddings.
% ~\textcolor{red}{Describe the architecture}.
% We map each image to a hidden vector using a five-layer convolution neural network with ReLU activation and batch normalization. The task embedding vector is mapped to the same space by fully connected layer. Then we use the element wise production between the feature as the input of the final linear logistic regression layer. The detailed information including the number of hidden neurons can be found in the supplementary. We tried following task embedding initializations:
% \begin{enumerate}[label=(\alph*)]

\vspace{-10pt}
% \small
\begin{enumerate}[itemsep=0em]
% [label=(\roman*)]
\item \textit{Label Embedding (Label Emb.)}: a binary length \(L\) vector \(\ev^{(t)}\) where \(e^{(t)}_\ell=1\) if and only if \(\ell\) is active for task \(t\). Prior knowledge of active tasks is required.
\vspace{-4pt}
\item \textit{Pseudo Label Embedding (Pseudo Emb.)}: we cluster training data with GMM with \(L\) components.
For each task \(t\), we calculate its empirical distribution on clusters as \(\ev^{(t)}\).\vspace{-4pt}
\item \textit{Random Initialization (Random Init.)}: initialize  \(\ev^{(t)}\) with  \(L\)-dimensional normal distributions.
\item \textit{Learned Embedding (Learned Emb.) (dim=\(M_0\))}: initialize task embeddings with \(M_0\) pre-selected tasks following the procedure in Section~\ref{sec:task_embeddings}.
\end{enumerate}
\vspace{-4pt}
\normalsize
% \normal
% \(i\)) 
% \(ii\)) \textit{Pseudo Label Embedding (Pseudo Emb.)}: we cluster training data with GMM with \(L\) components.
% For each task \(t\), we calculate its empirical distribution on clusters as \(\ev^{(t)}\).
% \(iii\)) \textit{Random Initialization (Random Init.)}: use \(L\)-dimensional normal distributions to random initialize \(\ev^{(t)}\).
% \(iv\)) \textit{Learned Embedding (Learned Emb.) (dim=\(M_0\))}: initialize task embeddings with \(M_0\) pre-selected tasks following the procedure in Section~\ref{sec:task_embeddings}.

\begin{figure}[t]
%   \begin{center}
% \scalebox{0.92}{
% \begin{minipage}[c]{0.45\textwidth}
        \includegraphics[width=0.48\textwidth]{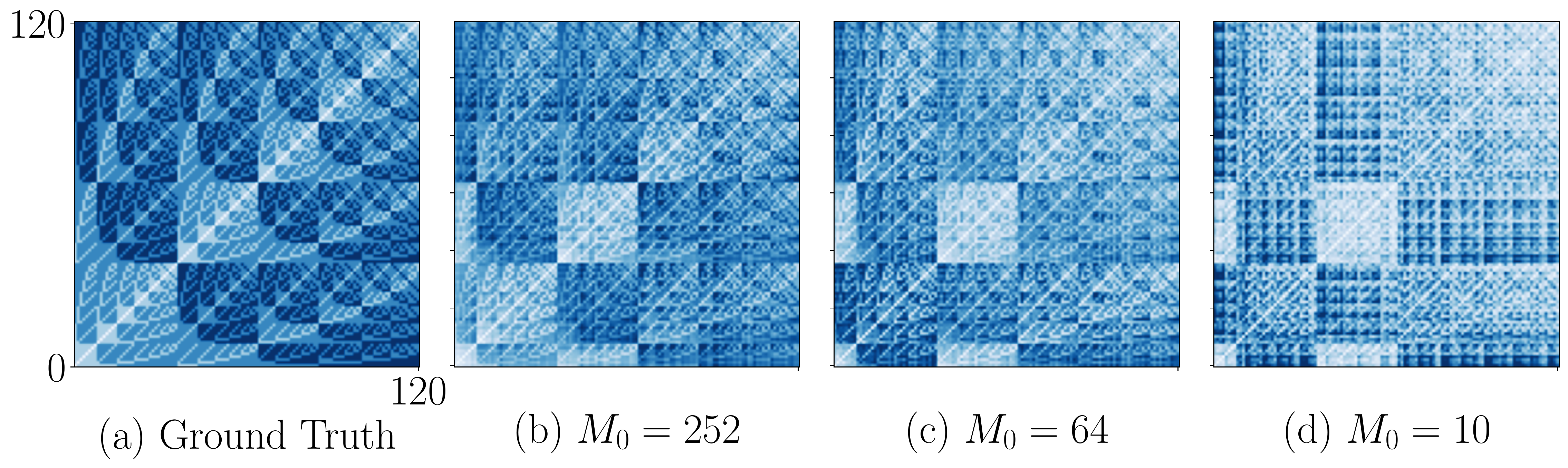}\vspace{-10pt}
    % \end{minipage}\hfill
    % \begin{minipage}[c]{0.43\textwidth}
        \caption{ Pairwise similarity between $120$ unseen tasks indexed by rows and columns. Lighter color represents stronger similarities. (a) Ground truth similarity measured by the number of overlapping active categories. (b-d) Cosine similarities measured by learned embeddings, by sampling (b) $M_0=252$, (c) $M_0=64$, (d) $M_0=10$ seed tasks.} \label{fig:embedding_showcase}
    % \end{minipage}
    %We train the multiple output heads with 252 tasks and use the outputs of MOH as the embedding of 3-active category case. Lighter color represents closer distance (a) ground-truth attribute distance (4 discrete values, 0,1,2,3) (b) $m = 252$ (c) $m = 64$ (d) $m = 10$
  \vspace{-6pt}
\end{figure}

We use the same architecture to integrate task embeddings, and use the training scheme for all initialization methods.
Details can be found in the %\textcolor{red}{Appendix}.
appendix.
%
% Put details in appendix.
% Following the training process in Figure~\ref{fig:occ_train_proc}, we use the multiple output head framework as the pre-trained model with a shared convolution neural network followed by ReLU activation and batch normalization go get the feature representation $h$. The output responses are $\vr=(r^{(1)},\ldots,r^{(M_0)})$, where $r^{(t)}=h^\top \phi^{(t)}$. $\ev^{(t)}$ is the empirical average over responses from task $t$. The number of tasks $M_0$ used for pre-training can be much smaller than the total number of tasks.
%
% \end{enumerate}

% \input{tables/mnist_auc_results}

% \input{tables/cifar10_embedding_results_v2}

% Put this part into the appendix.
% \input{figures/response_dist}

% Put this part into the appendix.
% For each method, the initial learning rate is \(1 \! \times \! 10^{-3}\) with decreases by 0.1 at epoch 200 and 250.
% We train 300 epochs in total and select the model with the highest AUC on the validation set as our final model.
% We train neural networks and fine-tune task embeddings.

As demonstrated in Table~\ref{table:toy_result} (left), CAD is generally harder when handling more tasks.
For two tasks \(t,t'\), label distribution of samples
% \(\|\ev^{(t)}-\ev^{(t')}\|\) 
% is larger when they share more active categories, which 
can capture the similarities of tasks precisely, leading to good performance of \textit{label embedding}.
\textit{Pseudo label embedding} tries to approximate label embedding, but it does not perform well when the number of tasks increases dramatically  due to the limitation of GMM.
\textit{Learned embedding} with a moderate number of seed tasks can do much better than GMM, and it achieves almost the same AUC as label embeddings.
% This is because learned embedding can better capture the high-level similarities among tasks, as will be further explained at the end of this section.
\textit{Random initialization} performs extremely poor even after fine tuning, indicting that task embedding initialization is critical in CAD.

% Label embedding use prior knowledge to perfectly capture task distance on the label space so it is not surprising to see label embedding works well. Dual embeddings with a moderate number of pre-training tasks can achieve almost the same AUC as label embeddings. Thus, one can expect that pre-training on a fraction of tasks generalize well to the remaining tasks. On the other side, pseudo label embedding can do relatively well when the number of tasks is small, but quickly fall behind dual embeddings when the number of tasks blows up. Random initialization is extremely bad even after fine tuning, indicting that embedding initialization is critical in MT-OCC\@.

\paragraph{Task generalization}
Surprisingly, the learned task embeddings can help identify unseen tasks fairly well without retraining or fine-tuning.
To demonstrate this, we use setting $k = m$ train the pre-embedding and conditional likelihood ratio estimation models.
Then we select another setting $k = n$ for testing.
% No data is shared between training and testing tasks.
For each testing task \(t\), its task embedding \(\ev^{(t)}\) comes from the \emph{existing} pre-embedding model.
We apply \(\ev^{(t)}\) to the \emph{existing} conditional likelihood ratio model for testing (active categories {\em vs}.\ the rest).

Table~\ref{table:toy_result} (right) shows the setup, where
\(m,n=2,3,4,5\) are selected as four exclusive sets of tasks (from easy to difficult) for training (indexed by each row) and testing (indexed by each column). We use (\(m\rightarrow n\)) to denote training on task set \(m\) and testing on task set \(n\). % \textcolor{red}{The embedding dimension for each task is 10.}
There is a generalization from training tasks to testing tasks since the testing performance is much better than random guessing (AUC = 0.5). The generalization can be attributed to two factors.
First, learned embeddings capture the distribution over new task well. This will be further discussed in the next paragraph.
Second, conditional likelihood ratio estimates can adapt to unseen task embeddings and make reasonable predictions. In general, it is harder to generalize from an easy set of tasks to a difficult set of tasks (\eg, \(2\!\rightarrow\!5\)) than vice versa (\(5\!\rightarrow\!2\)). This is because the easy set of tasks cannot precisely characterize the fine-grained correlations among the set of difficult tasks. 
However, this does not mean we can always train on hard tasks and expect to consistently achieve the best performance among all task sets since the result of \(3\!\rightarrow\!2\) is better than \(5\!\rightarrow\!2\), indicating the task set \(3\) is ``closer'' to task set \(2\) in the task semantic space, thus a better generalization performance.

% \textcolor{red}{In general, the model is easier to generalize to new tasks with fewer active categories \(k\).
% This is because the setting with larger \(k\) is harder to train because each task owns fewer samples, and pre-training on tasks with larger \(k\) are more informative due to richer task correlations, making it easier to generalize.}

% Put this into appendix.
% \paragraph{Dual embedding quality.}
% Dual embedding is the average over responses of a task. We demonstrate the response of an observation can well capture its attributes. We implement the log-likelihood response model with multiple output heads so each head $\vr^{(t)}(\xv)$ is the response of a task $t$, with $120$ tasks in total enumerating all combinations of $k=3$ active categories. In Figure~\ref{fig:response_cifar_09}, we show responses $\vr(\xv)$ for two images: one with a airplane and the other with a truck. Each image gets positive responses on heads containing its category and negative responses on the rest heads. We can easily discriminate each category by collaboratively using responses from all heads.

\paragraph{Visualize generalization through task embeddings}
The learned task embeddings are generalizable because they can measure high-level semantics, thus express correlations among different tasks.
We pre-train a task embedding model with all \(252\) tasks where \(k=5\) out of \(10\) categories are active.
Then we use the task embedding model to encode another \(120\) unseen tasks with \(k=3\) active categories.
We find that the learned task embeddings can approximate similarities between those \(120\) new tasks even the pre-trained network has not seen them before.
Figure~\ref{fig:embedding_showcase}(a) shows the ground-truth similarity, measured by the number of overlapping categories between two tasks.
It turns out the similarity are mainly preserved by learned embeddings in (Figure~\ref{fig:embedding_showcase}(b)).
Learned embeddings are robust, since even a few sub-sampled indices (64 indices in Figure~\ref{fig:embedding_showcase}(c) and 10 indices in Figure~\ref{fig:embedding_showcase}(d)) can approximately preserve task similarities.

% Dual embeddings are generalizable because they can measure similarity for unseen tasks. In the first showcase, we pre-train a response model with all $252$ tasks where $5$ out of $10$ categories are active. Then we test the dual embedding learned for another $120$ tasks with $3$ active categories. We find that dual embeddings can approximate similarities between those $120$ new tasks even the pre-trained network have not seen them before. Figure~\ref{fig:embedding_showcase}(a) shows the ground-truth similarity, measured by the number of overlapping categories between two tasks. It turns out the similarity are mainly preserved by dual embeddings in (Figure~~\ref{fig:embedding_showcase}(b)). Dual embeddings are robust, since even a few sub-sampled indices (64 indices in Figure~\ref{fig:embedding_showcase}(c) and 10 indices in Figure~\ref{fig:embedding_showcase}(d)) can approximately preserve task similarities.

% \input{experiment_likelihood_ratio}
\section{Conclusion and Future Work}\label{sec:conclusion}
In this paper, we studied a series of correlated anomaly detection problems with exposure data.
The key to jointly learn these AD tasks is to share information (collaborate) across them by explicitly or implicitly sharing data among tasks.
Our study finds that using conditional likelihood ratio estimation with the learned task embeddings can efficiently perform CAD on MovieLens 1M with over thousands of tasks.
We further studied the CIFAR10 dataset with a massive number of correlated yet different tasks, demonstrating that the proposed combination is superior to other possible choices, when data for each task is scarce, and its generalization ability.

The current work has a few limitations, which can be studied in future work.
For example, the choice of seed tasks can be more selective than random picking.
A full-fledged CAD may adapt to streaming data where new users and items come in real-time.
\bibliographystyle{apalike}
\bibliography{references}

\newpage
\begin{appendices}
% \begin{appendices}
\section{Log-likelihood Ratio Estimation}
% \subsection{}
% \begin{theorem}

% The output value of the logistic regression layer represents the the log-likelihood ratio between the positive and the negative samples.
% \]
% \end{theorem}

% \subsection{Proof of Section }
\subsection{Proof of Proposition 1 in Section 3.2.}
% In section 4, we present the optimal choice of the the base distribution $p(\xv)$. We restate below for convenience and give its proof.

\begin{proposition}\label{prop:kl}
    \(p(\xv)=\sum_t m^{(t)}q^{(t)}(\xv)\) is optimal for minimizing the expected KL divergence
    \begin{align}\label{eq:min_kl}
        \min_{p(\xv)\in\mathcal{P}}\mathbb{E}[\operatorname{KL}(q^{(t)}|p)] =\min_{p(\xv)\in\mathcal{P}}\sum_t\mathbb{E}_{ q^{(t)}(\xv)}\left[ m^{(t)} \log \frac{q^{(t)}(\xv)}{p(\xv)}\right].
    \end{align}
\end{proposition}

\begin{proof}
Expanding the expectation over tasks. We get
% The expectation is over tasks. so
\begin{align}
    \mathbb{E}[\operatorname{KL}(q^{(t)}|p)] = \sum_t m^{(t)} [\operatorname{KL}(q^{(t)}|p)].
\end{align}

We need to solve a minimization problem with constraints:
\begin{align}
     \min_{p(\xv)\in\mathcal{P}}\quad & \sum_t\mathbb{E}_{ q^{(t)}(\xv)}\left[ m^{(t)} \log \frac{q^{(t)}(\xv)}{p(\xv)}\right]\nonumber\\
     \text{s.t.}\quad & \int p(\xv) d \xv = 1.
\end{align}
With the Lagrange multiplier $\lambda$, we need to find function $p(\xv)$ to minimize the Lagrangian
% we need to find $p(\xv)$ to minimize
\begin{align}
    \sum_t\mathbb{E}_{ q^{(t)}(\xv)}\left[ m^{(t)} \log \frac{q^{(t)}(\xv)}{p(\xv)}\right] + \lambda\left(\int p(\xv) d\xv  - 1\right).
\end{align}
% The critical point is reached at
After removing additive terms unrelated to $p(\xv)$, we get
% Since we only care about the term related to $p(\xv)$
% and exchange the position of integral and summation and get
\begin{align}
    \int\bigg( \sum_t  -m^{(t)}q^{(t)}(\xv) \log p(\xv) + \lambda p(\xv) \bigg) d\xv.
\end{align}
% \begin{align}
%      \int\bigg(\sum_t -m^{(t)}q_t(\xv) \log p(\xv) + \lambda p(\xv) \bigg) d\xv
% \end{align}
% \begin{align}
    % \frac{1}{p(\xv)}\sum_t\mathbb{E}_{ q^{(t)}(\xv)}m^{(t)} + \lambda \bbone(supp(p(\xv))) = 0
% \end{align}
  In the calculus of variations, \(p(\xv)\) is varied by adding a function  \(\delta p(\xv)\) to it with \(\epsilon \) multiplier, where \(\epsilon \rightarrow 0\):
% Take the variation  over .
\begin{align}
    \int\bigg(\sum_t -m^{(t)} q^{(t)}(\xv) \log \big(p(\xv) + \epsilon \delta p(\xv)\big) + \lambda (p(\xv)+ \epsilon \delta p(\xv)) \bigg) d\xv .
\end{align}
The change in the value to first order in $\epsilon$ should be zero. By taking derivative w.r.t. $\epsilon$ and let $\epsilon = 0$, we get
\begin{align}
    \int\bigg( \sum_t  -m^{(t)}q^{(t)}(\xv) \frac{\delta p(\xv)}{p(\xv)}  + \lambda (\delta p(\xv)) \bigg) d\xv = 0.
\end{align}
% yes！！！

% So we can get
The above equation is valid for any function $\delta p(\xv)$
\begin{align}
    \lambda -  \frac{1}{p(\xv)}\sum_t m^{(t)}q^{(t)}(\xv) = 0.
\end{align}

% \begin{align}
    % L =
% \end{align}

% \begin{align}
% \delta J &= \int_a^b \left( \frac{\partial L}{\partial f} \delta f(x) + \frac{\partial L}{\partial f'} \frac{d}{dx} \delta f(x) \right) \, dx \, \\
% &= \int_a^b \left( \frac{\partial L}{\partial f} - \frac{d}{dx} \frac{\partial L}{\partial f'} \right) \delta f(x) \, dx \, \\
% & + \, \frac{\partial L}{\partial f'} (b) \delta f(b) \, - \, \frac{\partial L}{\partial f'} (a) \delta f(a) \,
% \end{align}
% We need optimize the function

% If we take the variational derivative of $p(\xv)$ for part in the integral,

After reordering terms, we get
\begin{align}
\label{eq:final}
    p(\xv) = \frac{1}{\lambda}\sum_t m^{(t)} q^{(t)}(\xv).
\end{align}

Since $\int p(\xv) = 1$ and $\sum_t m^{(t)} = 1$, taking the integral over $\xv$ on both sides of Equation \ref{eq:final} permits $\lambda = 1$. Finally, we get
\begin{align}
    p(\xv) = \sum_t m^{(t)} q^{(t)}(\xv).
\end{align}
% \begin{align}
    % p(\xv)\bbone_{supp(p(\xv))} & = \frac{1}{\lambda} \sum_t\mathbb{E}_{ q^{(t)}(\xv)}m^{(t)} \\
    % & = \frac{1}{\lambda} \sum_t m^{(t)} = \frac{1}{\lambda}
% \end{align}
% If we take the derivative over $p(\xv)$, we arrive the critical point at
% \begin{align}
This concludes our proof.
\end{proof}

% \subsection{Proof of Section }
\subsection{Proof of the optimal solution to Equation 3}
Suppose we have two distributions \(q(\xv)\) and $p(\xv)$, their log-likelihood ratio $r(\xv)=\log(q(\xv)/p(\xv))$ can be estimated as follows.

First, we independently draw $N$ samples from $q(\xv)$ and label those data as positive: $\{\xv_i,y_i=1\}_{i=1}^N$. Similarly we draw $N$ samples from $p(\xv)$ as label those as negative: $\{\xv_i,y_i=-1\}_{i=N+1}^{2N}$. We combine all positive samples and negative samples into one dataset $D=\{\xv_i,y_i\}_{i=1}^{2N}$. In the limit of $N\rightarrow\infty$, the mixed distribution satisfies:
\begin{align}
p(y=1|\xv)=\frac{q(\xv)}{p(\xv)+q(\xv)}.
\end{align}
Then we train a binary classifier on the mixed dataset $D$. Suppose we are using a linear logistic regression model:
% Given a dataset $\{ \mathbf{x}_i , y_i \}_{i=1}^n$, $y_i \in \{\pm1\}$ the logistic regression model is defined by
\begin{align}
\hat{p}(y\,|\,\mathbf{x};\,\bm\theta) = \sigma(y\cdot\bm\theta^\top\mathbf{x}),
\end{align}
where $\sigma$ is the sigmoid function:
\begin{align}
\sigma(z) = \frac{1}{1+e^{-z}}.
\end{align}
Then
\begin{align}
\log\frac{q(\xv)}{p(\xv)}&=\log\frac{p(y=1|\xv)}{1-p(y=1|\xv)}\nonumber\\
&\approx\log\frac{\hat{p}(y=1|\xv;\,\bm\theta)}{1-\hat{p}(y=1|\xv;\,\bm\theta)}\nonumber\\
&=\log \frac{\sigma(\bm\theta^\top\mathbf{x})}{\sigma(-\bm\theta^\top\xv)}\nonumber\\
&=\bm\theta^\top\mathbf{x}.
  % \log\frac{p(y=+1\,|\,\mathbf{x};\,\bm\theta)}{p(y=-1\,|\,\mathbf{x};\,\bm\theta)}
  % =\log \frac{\sigma(\bm\theta^\top\mathbf{x})}{\sigma(-\bm\theta^\top\xv)} = \bm\theta^\top\mathbf{x}.
  % & = \log \frac{1+e^{\thetav^\top\xv}}{1+el^{-\thetav^\top\xv}} \\
  % & = \log  e^{\thetav^\top\xv} \frac{1+e^{\thetav^\top\xv}}{e^{\thetav^\top\xv}+1} \\
  % & = \log  e^{\thetav^\top\xv} \\
  % & = \bm\theta^\top\mathbf{x}
\end{align}
One can use a non-linear model implemented with a neural network $f(\xv,\bm\theta)$ to get a better estimation of the log-likelihood ratio.

% \section{Conditional Normalizing Flow Model}

% \input{sub_nf}
% As we wrote in the  \textcolor{red}{Equation (2)}, we added the task information in two different ways, the prior and weight masks. 

\section{Conditional Normalizing Flow}
\subsection{Feature extractors.}
% Learning a good image generator with normalizing flow can not guarantee obtaining a good out of distribution detector (\citet{kirichenko2020normalizing}), since the pixel correlations could be more important than semantics when we want to generate some high-quality pictures. Thus the details of the content are less helpful when we use ``category'' as the OOD detection criterion. In the meantime, a good normalizing flow model designed for image generation often requires large amounts of parameters and training time, which is hard to tune. It is also not applicable when the number of tasks is large, or when each task only owns a few samples. 

A good image generator implemented with normalizing flows is not necessary a good out-of-distribution detector, since normalizing flows tend to capture pixel correlations rather than semantics during training~\citep{kirichenko2020normalizing}. Moreover, normalizing flow models for image generation often requires a large number of parameters and long training time, which is not-applicable when the number of tasks is large. 

% To avoid the above problems, we use extracted low dimensional features instead of pixel-based raw images. The features are extracted using unsupervised contrastive learning (\citet{chen2020simple}) with a Resnet18 as the backbone. We test the feature quality with a linear classifier and get 90.2\% accuracy. This unsupervised learned feature extractor guarantees that the label information is not leaked while preserving more content information, comparing to those feature extractors trained by supervised classification.

To resolve above issues, we use pre-trained low-dimensional features instead of raw images to train normalizing flows. The features are extracted using unsupervised contrastive learning~\citep{chen2020simple} with a ResNet-18 as the backbone model. The training accuracy on the extracted features is 90.2\% with a linear classifier, demonstrating high feature quality. The unsupervised learning procedure guarantees that the extracted feature does not leak its label information while preserving image semantics.

\subsection{Two ways to implement conditional normalizing flows.}
% In the following equation, we summarize all task related thing with \(\ev^{(t)}\) for  brevity. Actually it is added to the prior distribution \(\pi(\xv)\) and the transformation \(f\) in two different ways. 
In the following, we use \(\ev^{(t)}\) to represent task-specific information, which can involve in the prior \(\pi\) and the invertible mapping \(f\) as follows:
% We will give a more detailed explanation below.
\[
    L = -\log q(\xv|\ev^{(t)}) = -\log\pi_{\ev^{(t)}}(f^{-1}_{\ev^{(t)}}(\xv)) - \log\left|\det\frac{\partial f^{-1}_{\ev^{(t)}}(\xv)}{\partial\xv}\right|.
\]
\begin{paragraph}{Task-specific priors}
% The prior distribution of normalizing flow \(\sim \pi(\xv)\) is usually standard multivariate normal distribution with zero means and identity covariance.
For un-conditional normalizing flows, the prior $\pi$ is usually set as a centered isotropic Gaussian. 
% In our setting, each task \(t\) has its prior Gaussian distribution \(\pi_{\ev_{prior}^{(t)}}\) with a uniformly sampled mean vector and identity covariance. Note that the mean is sampled initially and fixed during the training process for each task.
In our conditional case, each task \(t\) has its task-specific Gaussian prior \(\pi_{\ev_{prior}^{(t)}}\) with a uniformly sampled mean vector and an identity covariance matrix.
\end{paragraph} 

\begin{paragraph}{Task-specific mappings}
  MAF uses Masked Autoencoder for Distribution Estimation (MADE) as its building blocks. The core idea behind MADE is autoregressive flow, which generates data recursively by 
\[
    x_i = u_i \exp \alpha_i + \mu_i 
\]
% Where \(\mu_i = f(\xv_{1:i-1}),~\alpha_i = g(\xv_{1:i-1}) \). \(\uv\)  is the random distribution generated from the previous layer or the prior distribution. In our MADE implementation, function \(f\) and \(g\) are both 4-layer autoregressive fully connected layer with masking. The task specific features are combined with the output of each MADE layer by scaling and biasing, i.e. \(f_{\ev^{(t)}} = f \odot \textit{tanh}(\ev_{scale}^{(t)}) + \textit{tanh}(\ev_{bias}^{(t)})\). The hyperbolic tangent function is used to stabilize the training, and it can be replaced with deep neural networks. \(\ev_{scale}^{(t)}\) and \(\ev_{bias}^{(t)}\) are trainable task embeddings.
Where \(\mu_i = f(\xv_{1:i-1}),~\alpha_i = g(\xv_{1:i-1}) \). \(\uv\) is randomly generated from the previous layer or the prior distribution. In our MADE implementation, function \(f\) and \(g\) are both 4-layer autoregressive fully connected layer with masking. The task specific features are combined with the output of each MADE layer by scaling and biasing, i.e. \(f_{\ev^{(t)}} = f \odot \textit{tanh}(\ev_{scale}^{(t)}) + \textit{tanh}(\ev_{bias}^{(t)})\). The hyperbolic tangent function is used to stabilize the training, and it can be replaced with deep neural networks. \(\ev_{scale}^{(t)}\) and \(\ev_{bias}^{(t)}\) are trainable task embeddings.
% The \(\textit{NN}\) is implemented with a single \(\textit{tanh}\) activation function in order to stabilize the training.
These parameters are shared among different \(i\) in one layer. The same technique also applies to function \(g\). For more details on MAF, please refer to the origin paper \citet{germain2015made,papamakarios2017masked}.
\end{paragraph}

% The other way is adding task-specific parameters. 
% The condition is added to the prior of each task  as we have shown in the loss function 
% The task information is 
 Our implementation refers to two Github codebases \footnote{pytorch-normalizing-flows:  \url{https://github.com/karpathy/pytorch-normalizing-flows}.}$^{,}$\footnote{pytorch-flows:  \url{https://github.com/ikostrikov/pytorch-flows}}.

\section{Real scenarios for CAD}

% \section{Hyper-parameter Selection in experiment section}
\section{Training Details}
In this section, we give full descriptions of models in our experiments.

% In the experiment section, we did not give a full description of the detail of each model. This section serves as its supplementary.

\subsection{MNIST and CIFAR10}

\begin{figure}[htb!]
\centering
    \includegraphics[width=.4\textwidth]{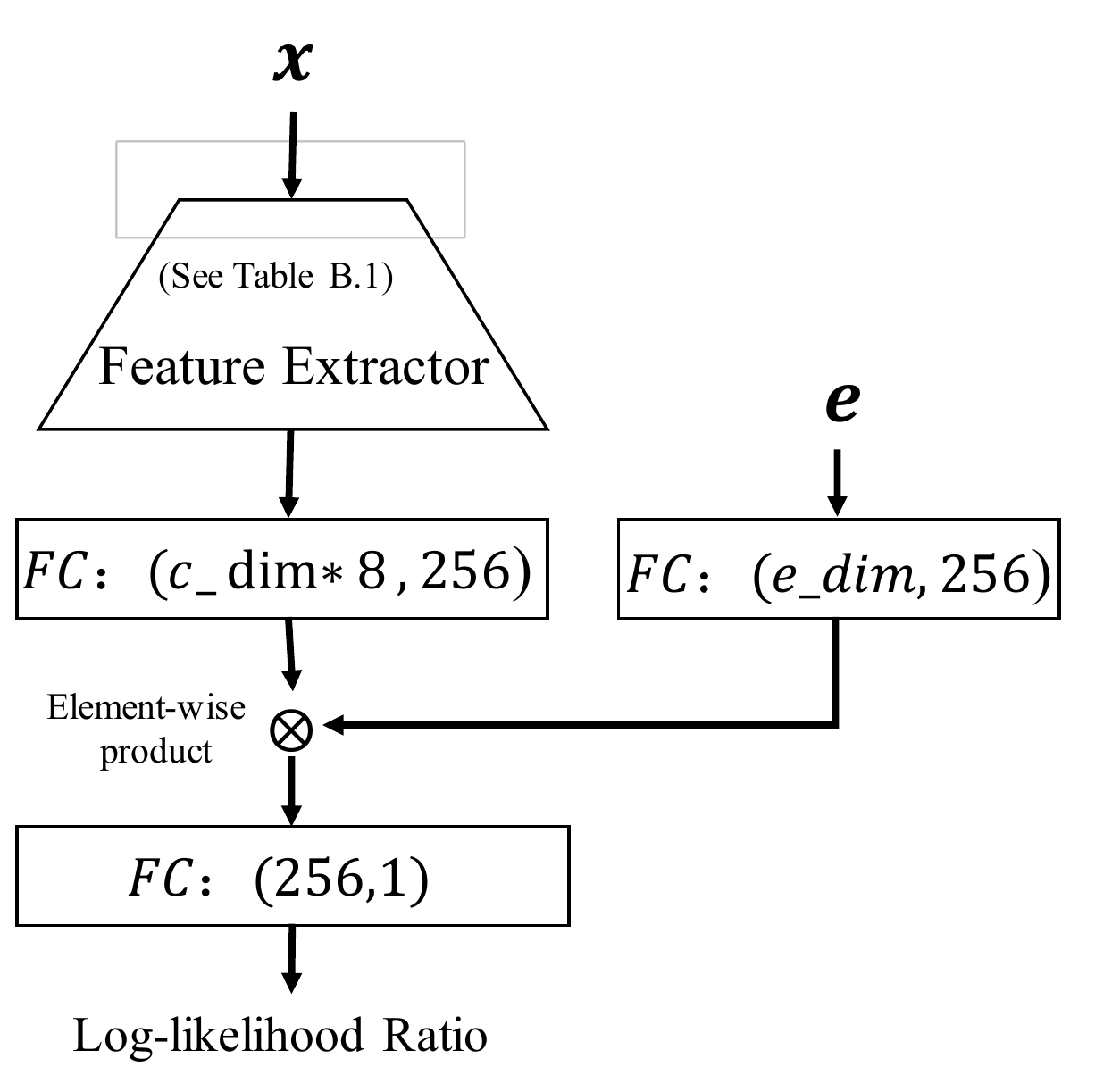}\vspace{-10pt}
    \caption{The framework we use to train MNIST and CIFAR10. $\xv$ denotes the input matrix. $\ev$ is the task embedding. FC represents the fully connected layer. The brackets followed by contains the dimension of its input and output. e\_dim is the dimension of the task embedding. The definition of c\_dim and the detail of feature extractor can be found in Table \ref{table:appendix_1}.}
\label{fig:appendix_1}
%\end{figure}
\bigskip
%\begin{figure}[]
%\centering
    \includegraphics[width=.3\textwidth]{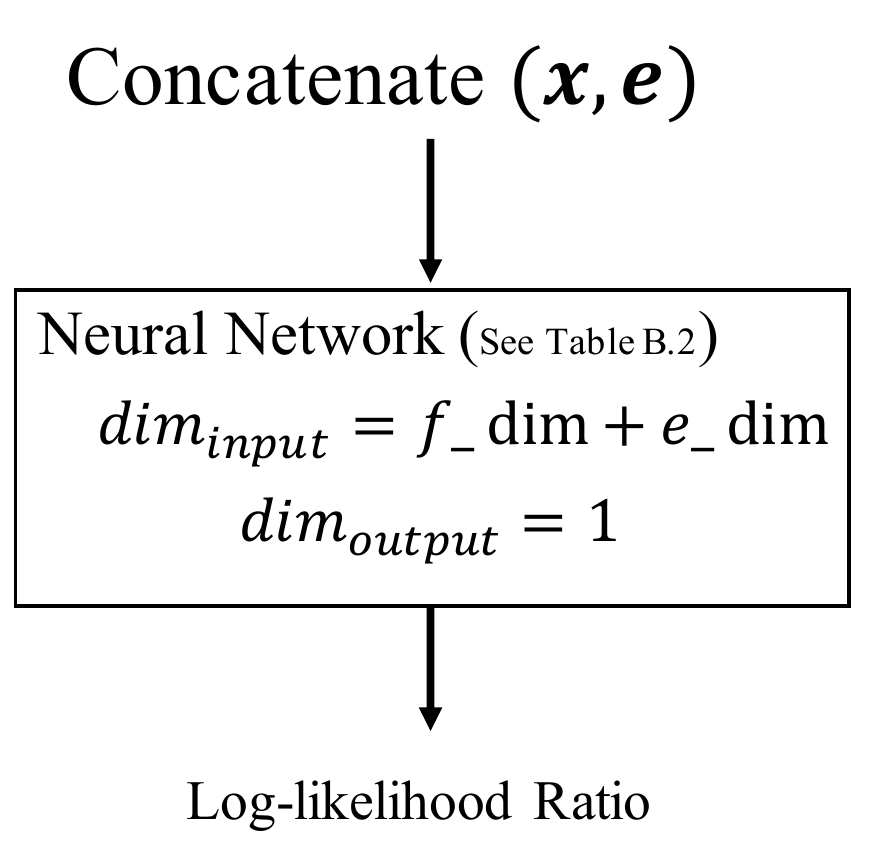}\vspace{-10pt}
    \caption{The framework we use to train MovieLens 1M. The definition of f\_dim, e\_dim and the detail of neural network can be found in Table  \ref{table:appendix_2}.}
\label{fig:appendix_2}
\end{figure}

Figure \ref{fig:appendix_1} demonstrates the network structure of the likelihood ratio estimation model ($f(\xv,\ev,\bm\theta)$ in  Algorithm 1)
% of the network structure for Algorithm 1
for MNIST and CIFAR10. We resize all MNIST figures from their original size $28\times28$ to $32\times32$ to align with CIFAR10 figure sizes. The data augmentations include random crop with padding (Pad the image to \(40\times40\) with zeros, then randomly choice a \(32\times32\) area) and random horizontal flip. We normalize pixels for each channel such that their mean equals to zero and variance equals to one.

% The pre-embedding model uses a similar feature extractor, followed by a fully connected layer mapping from feature dimension c\_dim*8 to 256. Then $M_0$ fully connected output heads are used to encode responses.
The pre-embedding model uses a similar feature extractor, followed by a fully connected layer mapping from feature dimension c\_dim*8 to 256. Then $M_0$ fully connected output heads are used to predict the likelihood ratio for $M_0$ seed tasks.
%Except the lack of embedding vector $\ev$, the only difference is that it owns $M_0$ fully connected output heads.
% don't know if this is necessary ->
% These two frameworks look very similar. Because element-wise production between the output feature and the mapped embedding vector behaves the same as element-wise production between the parameters of output heads and the mapped embedding vector.

\subsection{MovieLens 1M}
\subsubsection{Feature Extractor}
% Unlike CIFAR10 and MNIST where each image has its raw pixel representation, the user feature is implicitly represented by the interaction between movies and users. We treat MovieLens as a click-through rating prediction dataset and use a deep factorization-machine~\citep{guo2017deepfm} to extract user embeddings \(\xv\). We use the embedding layer after the training of the factorization model to extract features of each user. Since we are working on a item-recommendation problem where the movies are recommended to the users, all user are visible to the model during the training process. We can get the features of each user. We implement this method using code from \citet{cao2019unifying}.
Unlike CIFAR10 and MNIST where each image has its raw pixel representation, the user feature is implicitly represented by the interaction between movies and users. We treat MovieLens as a click-through rating prediction dataset and use a deep factorization-machine~\citep{guo2017deepfm} to extract user embeddings \(\xv\). After training the factorization model, we use the trained embedding layer to extract features of each user.
During the training process, all users are visible to the model, so we can get the feature of each user. We implement this method using code from \citet{cao2019unifying}.

\subsubsection{Hyper-parameters}
% Figure \ref{fig:appendix_2} shows our network structure  ($f(\xv,\ev,\bm\theta)$ in  Algorithm 1) on MovieLens 1M dataset. The pre-embedding model uses the same network structure. We remove embeddings from the input and add $M_0$ output heads.
Figure \ref{fig:appendix_2} demonstrates the network structure of the likelihood ratio estimation model ($f(\xv,\ev,\bm\theta)$ in  Algorithm 1) for MovieLens 1M dataset. The pre-embedding model uses the same network structure. The only difference is that we change the way to integrate task embeddings from adding them to the input to adding $M_0$ output heads in the pre-embedding model.
% with two differences. One is the dimension of embedding is zero, the other is the dimension of output is $M_0$.

% \begin{table}[!tb]
% \caption{Performance with/without the stop-gradient trick on CIFAR10 in terms of AUC\;(\(\times\)100).}\label{table:cifar_sg}

% \begin{tabular}{|c|c|c|c|}
% \hline
% %
% with stop-gradient\textbackslash tasks & 2 (45) & 3 (120) & 5 (252) \\ \hline
% Yes                                     & 95.97 & 93.07 & 78.96  \\ \hline
% No                                      & 75.73 &  & 52.88  \\ \hline
% \end{tabular}
% \end{table}

% \section{Multitask Embedding}

\begin{table}[h]
\centering
\caption{The network structure of feature extractor for CIFAR10 and MNIST. Each operation contains four parts: (kernel size, stride, padding size), number of output channel, batch normalization (BN) and activation function. In our setting, we use c\_dim = 64. The output size is a three-dimensional vector: (width, height, channel). The input channel is 3 for CIFAR10  and 1 for MNIST.}

\begin{tabular}{|c|l|c|}
\hline
Stage & \multicolumn{1}{c|}{Operation}              & Output Size  \\ \hline
Input &   \multicolumn{1}{c|}{Data Augmentation}       & (32,32,3 or 1)    \\ \hline
1     & (4, 2, 1), c\_dim, BN, ReLU  & (16, 16, c\_dim) \\ \hline
2     & (4, 2, 1), c\_dim*2, BN, ReLU  & (8, 8, c\_dim*2)   \\ \hline
3     & (5, 1, 0), c\_dim*4, BN, ReLU & (4, 4, c\_dim*4)  \\ \hline
4     & (4, 1, 0), c\_dim*8, BN,       & (1,1, c\_dim*8)   \\ \hline
\end{tabular}
\label{table:appendix_1}
\end{table}

%\begin{figure}[]
%\centering
%    \includegraphics[width=.3\textwidth]{figures/movielens_hyp_framework.pdf}
%    \caption{The framework we use to train Movielens-1M. The definition of f\_dim, e\_dim and the detail of neural network can be found in Table  \ref{table:appendix_2}.}
%\label{fig:appendix_2}
%\end{figure}

\begin{table}[h]
\centering
\caption{The network structure of logistic regression for MovieLens 1M. Each operation contains three parts: (input size, output size) of fully connected layer, activation function and dropout rate. f\_dim represents the dimension of input data. e\_dim is the dimension of the task embedding. We use f\_dim=100, e\_dim = 16.}
\begin{tabular}{|c|l|}
\hline
Stage & \multicolumn{1}{c|}{Operation}                   \\ \hline
% Input & \multicolumn{1}{c|}{Data Augmentation of $\xv$}      \\ \hline
1     & (f\_dim + e\_dim, 32), ReLU, Dropout (0.5)              \\ \hline
2     & (32, 32), ReLU, Dropout (0.5)                            \\ \hline
3     & (32, 16), ReLU, Dropout (0.3)                            \\ \hline
4     & (16, 1), Linear , No Dropout                                               \\ \hline
\end{tabular}
\label{table:appendix_2}
\end{table}

\section{More results on MNIST}
Similar as Table 3 (left) in the main text for the CIFAR10 dataset, we study different task embedding initialization methods (see Section 5.2 in the main text for definitions of those task embeddings) on the MNIST dataset with two CAD settings where $k=4$ with $C_{10}^4=210$ tasks, and $k=5$ with $C_{10}^5=252$ tasks. We skip the setting with $k=2$ and $k=3$ since those settings are too simple such that every method we tried can get almost $100\%$ AUC\@. The results is shown in Table \ref{table:toy_result_mnist}.

\begin{table*}[h]%\footnotesize
  \caption{%\footnotesize
    %   Cross-domain matching results in terms of Recall@$K$ (R@K). The results in the first section is on Flickr30K. The results in the second section is on MSCOCO.
    Testing AUC\;(\(\times\)100) on MNIST.  Label embedding uses prior knowledge (ground truth label) so it has a natural advantage comparing with other methods (Note that the first method ``label embedding'' uses extra knowledge and is marked with *. The best results without extra knowledge are shown in bold.).
    %*Means extra knowledge (ground truth category) is used in training.
  }\label{table:toy_result_mnist}
  \begin{center}
    \begin{small}
      \begin{sc}
        % \vspace{-3mm}
        % \begin{tabular}{lccccccc}
        \begin{tabular}{ccc}
          \toprule
        %   \multicolumn{1}{c}{}                       & \multicolumn{2}{c}{MNIST}                                                                            \\
          \multicolumn{1}{c|}{Embedding Init.~\textbackslash~K (\#tasks)}            & 4 (210)                    & 5 (252)      \\
          \midrule
          \multicolumn{1}{c|}{Label Embedding*}       & 99.88                      &99.67      \\
          \multicolumn{1}{c|}{Pseudo Label Embedding} & 97.88                      & 98.67  \\
          %\multicolumn{1}{c|}{Multiple Output Heads}  & \textbf{99.94}             & \multicolumn{1}{c|}{99.76}          & \textbf{96.67} & \textbf{95.08} & \textbf{93.54} & 89.38         \\
          \multicolumn{1}{c|}{Random Initialization}  & 97.02                      & 86.80 \\
          \multicolumn{1}{c|}{Learned Embedding ($M_0 = 10$)}        & 98.88                      & 99.67      \\
          \multicolumn{1}{c|}{Learned Embedding ($M_0 = 64$)}        & \textbf{99.93}             & \textbf{99.90} \\
          % \multicolumn{1}{l}{}                       & \multicolumn{1}{l}{} & \multicolumn{1}{l}{}         & \multicolumn{1}{l}{} & \multicolumn{1}{l}{} & \multicolumn{1}{l}{} & \multicolumn{1}{l}{} \\ \hline
          \bottomrule
        \end{tabular}
      \end{sc}
    \end{small}
    % \vspace{-5mm}
  \end{center}
\end{table*}
% \end{appendices}

\end{appendices}
\iffalse
References follow the acknowledgements.  Use an unnumbered third level
heading for the references section.  Please use the same font
size for references as for the body of the paper---remember that
references do not count against your page length total.

\fi

\end{document}